\documentclass[iicol, pdflatex, sn-mathphys-num]{sn-jnl}

\usepackage{graphicx}%
\usepackage{multirow}%
\usepackage{amsmath,amssymb,amsfonts}%
\usepackage{amsthm}%
\usepackage{mathrsfs}%
\usepackage[title]{appendix}%
\usepackage{xcolor}%
\usepackage{textcomp}%
\usepackage{manyfoot}%
\usepackage{booktabs}%
\usepackage{algorithm}%
\usepackage{algorithmicx}%
\usepackage{algpseudocode}%
\usepackage{listings}%

\usepackage{soul}
\usepackage{xcolor}
\usepackage{mathtools}
\usepackage{tabularx}
\usepackage{stfloats}
\usepackage{array,booktabs,makecell}
\usepackage{arydshln}
\usepackage{orcidlink}

\newcommand{\dottedmidrule}{\hdashline[0.3pt/2pt]}

\newcolumntype{N}{>{\centering\arraybackslash}m{1.5cm}} 
\newcolumntype{Y}{>{\centering\arraybackslash}m{2.5cm}}  
\newcolumntype{L}{>{\raggedright\arraybackslash}X}        

\theoremstyle{thmstyleone}%
\theoremstyle{thmstyletwo}%

\theoremstyle{thmstylethree}%

\begin{document}

\title[Probabilistic Human Intent Prediction for Mobile Manipulation: An Evaluation with Human-Inspired Constraints]{Probabilistic Human Intent Prediction for Mobile Manipulation: An Evaluation with Human-Inspired Constraints}



\author*[1]{\fnm{Cesar Alan} \sur{Contreras}\orcidlink{0009-0005-2866-8672}}\email{cac214@bham.ac.uk}

\author*[2]{\fnm{Manolis} \sur{Chiou}\orcidlink{0000-0002-9779-4067}}\email{m.chiou@qmul.ac.uk}

\author*[1,3]{\fnm{Alireza} \sur{Rastegarpanah}\orcidlink{0000-0003-4264-6857}}\email{a.rastegarpanah@aston.ac.uk}

\author[4]{\fnm{Michal} \sur{Szulik}}\email{michal.szulik@uknnl.com}

\author[1]{\fnm{Rustam} \sur{Stolkin}\orcidlink{0000-0002-0890-8836}}\email{r.stolkin@bham.ac.uk}


\affil[1]{\orgdiv{School of Metallurgy \& Materials},
          \orgname{University of Birmingham},
          \orgaddress{\city{Birmingham, B152SE}, \country{United Kingdom}}}

\affil[2]{\orgdiv{School of Electronic Engineering and Computer Science},
          \orgname{Queen Mary University of London},
          \orgaddress{\city{London, E14NS}, \country{United Kingdom}}}

\affil[3]{\orgdiv{School of Computer Science and Digital Technologies},
          \orgname{Aston University},
          \orgaddress{\city{Birmingham, B47ET}, \country{United Kingdom}}}

\affil[4]{\orgname{United Kingdom National Nuclear Laboratory Ltd.},
          \orgaddress{\city{Warrington, WA36AE}, \country{United Kingdom}}}


\abstract{Accurate inference of human intent enables human-robot collaboration without constraining human control or causing conflicts between humans and robots. We present GUIDER (Global User Intent Dual-phase Estimation for Robots), a probabilistic framework that enables a robot to estimate the intent of human operators. GUIDER maintains two coupled belief layers, one tracking navigation goals and the other manipulation goals. In the Navigation phase, a Synergy Map blends controller velocity with an occupancy grid to rank interaction areas. Upon arrival at a goal, an autonomous multi-view scan builds a local 3D cloud. The Manipulation phase combines U$^{2}$-Net saliency, FastSAM instance saliency, and three geometric grasp-feasibility tests, with an end-effector kinematics-aware update rule that evolves object probabilities in real-time. GUIDER can recognize areas and objects of intent without predefined goals. We evaluated GUIDER on 25 trials (five participants × five task variants) in Isaac Sim, and compared it with two baselines, one for navigation and one for manipulation. Across the 25 trials, GUIDER achieved a median stability of 93–100\% during navigation, compared with 60–100\% for the BOIR baseline, with an improvement of 39.5\% in a redirection scenario (T5). During manipulation, stability reached 94–100\% (versus 69-100\% for Trajectron), with a 31.4\% difference in a redirection task (T3). In geometry-constrained trials (manipulation), GUIDER recognized the object intent three times earlier than Trajectron (median remaining time to confident prediction 23.6 s vs 7.8 s). These results validate our dual-phase framework and show improvements in intent inference in both phases of mobile manipulation tasks.}




\maketitle

\section{Introduction}\label{Introductiomn}

The capacity for a robot to infer and predict human intent is a key capability for enabling fluent human-robot collaboration in unstructured and dynamic environments \cite{contreras2025mini, shao2024constraint, li2022review}.  Although certain mobile-manipulation tasks can be performed without direct collaboration, this ability becomes especially valuable when human-robot teaming is required.  Mobile manipulators, which combine the locomotive flexibility of a mobile base with the fine-motor dexterity of a manipulator arm \cite{contreras2025mini, chiou2016experimental}, hold broad potential in industrial automation, service robotics, medical assistance, assistive technologies, search and rescue, and decommissioning operations \cite{contreras2025mini, gopinath2020active}.

Accurate and fast intent prediction can improve task efficiency and completion rates by minimising unnecessary actions and delays \cite{oh2021system, luo2022human, wu2021spatial}. Like in any human team, a human operator and a robot must first align their goals. A robot that fails to recognise the operator's intent can lead to friction, or outright conflicting behaviours \cite{negotiation2022}. On the other hand, when the robot understands human intent, it can help reduce the cognitive burden on the human operator by allowing algorithms to solve low-level motion decisions \cite{ramesh2021robot, chiou2022robot, contreras2025mini}, while also raising safety by lowering the risk of misinterpretations or delayed responses \cite{panagopoulos2021bayesian, shao2024constraint, javdani2018shared, ding2011human}.  Effective human-robot interaction (HRI), therefore, relies on the ability of the robot to follow immediate commands while also reasoning about human goals \cite{panagopoulos2021bayesian, gopinath2020active, tahboub2006intelligent, li2022review, song2024robot}. With complementary capabilities, each partner can tackle sub-tasks that are difficult for the other.

However, a significant limitation of many existing intent recognition frameworks is their focus on isolated subsystems where navigation and manipulation are studied as two separate problems, each with its own predictor and little or no information flow between them \cite{contreras2025mini, panagopoulos2021bayesian, oh2021system, shao2024constraint, chiou2016experimental, luo2022human}.  Such compartmentalization produces subsystems that excel at one aspect of the task yet fail in complex, multimodal scenarios that blend navigation and manipulation.  

Specialized predictors perform well on narrowly defined problems, but they seldom cope with the diverse and rapidly changing intentions encountered in real-world collaboration.  A predictor tuned for a simple pick-and-place sequence, for instance, can fail during a multi-step collection routine that includes navigating between workstations, switching tools, manipulating different objects, and loosely specified objectives. Practical tasks require integrated reasoning. A robot must recognize that the goal of the operator may shift between navigation and manipulation \cite{salzmann2020trajectron, contreras2025mini}, and that the robot’s physical capabilities bound each modality.

To address these issues, we propose GUIDER (Global User Intent Dual-phase Estimation for Robots), a unified probabilistic framework that maintains coupled belief layers for navigation intent and end-effector intent.  These layers fuse controller motion, map context, visual saliency, segmentation, and grasp-feasibility cues, enabling the system to update area-level and object-level hypotheses continuously.

\begin{figure}[!b]
    \centering
    \includegraphics[width=1.0\linewidth]{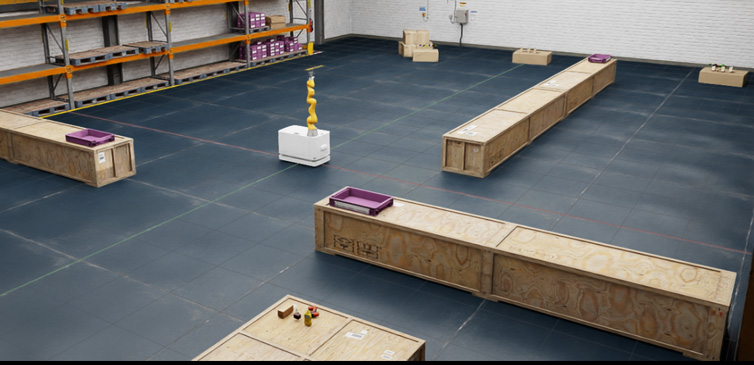}
    \caption{The experimental environment prepared in the Isaac Sim simulator with the Kuka KMR IIWA mobile manipulator shown in the warehouse setting.}
    \label{fig:simulated_environment}
\end{figure}

\begin{figure*}[ht]
    \centering
    \includegraphics[width=1\linewidth]{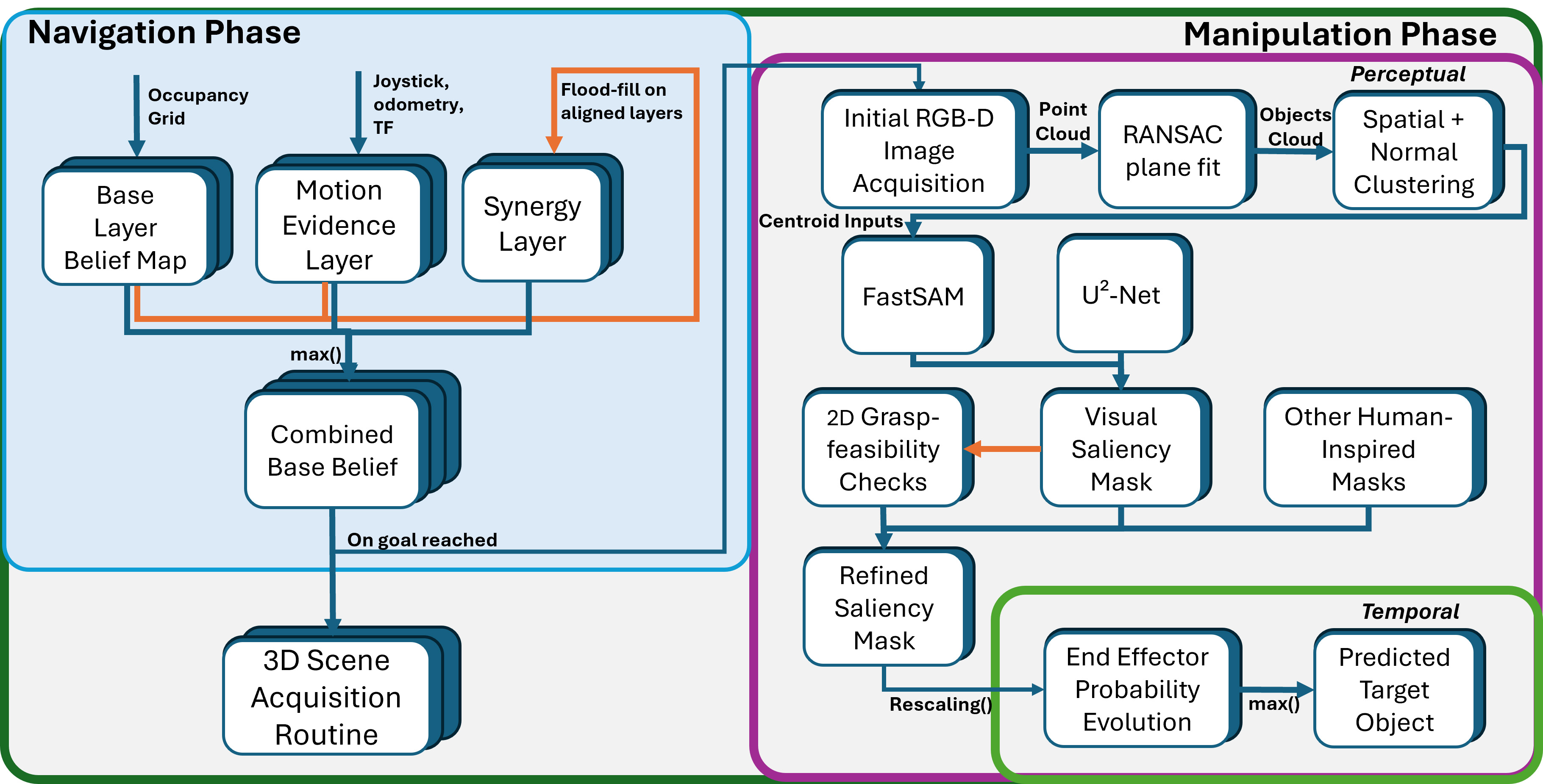}
    \caption{The architecture of GUIDER, the proposed probabilistic intent inference pipeline. The Navigation Phase combines a map-based Base Layer with a Motion Evidence Layer, creating a Synergy Layer through a flood-fill algorithm. The merging and maximum of these layers predict the target interaction area, utilizing a combined base belief. Upon arrival at a navigation goal, the Manipulation Phase is triggered, with simultaneous 3D scene acquisition to aid the operator's spatial awareness. The manipulation phase performs 3D geometric analysis (RANSAC, HDBSCAN clustering) to find object centroids, which are used to prompt FastSAM for 2D instance segmentation. The resulting masks are fused with a U²-Net saliency mask, 2D Grasp-feasibility Checks, and other human-inspired constraints to create a Refined Saliency Mask. Finally, an End Effector Probability Evolution model updates these probabilities in real time using manipulator kinematics to yield the final Predicted Target Object.}
    \label{fig:intent_pipeline}
\end{figure*}

GUIDER operates in two phases. In the navigation phase, the system updates the area-level intent. A synergy map blends controller inputs with the map, diffusing probability toward cells that lie ahead of the robot; its peak indicates \emph{where} the operator is likely to interact next. When that region is reached, a 3D scan is performed. Then, the manipulation phase can activate, converting depth, visual saliency, segmentation, and multiple grasp-affordance checks into object-level hypotheses, which tell the robot \textit{what} the human intent is and \emph{why} it is feasible with its tool. The operator remains free to move the base or arm at any time, and both belief layers continuously update while sharing information downstream. Figure \ref{fig:simulated_environment} depicts the warehouse environment used in the paper, while Figure \ref{fig:intent_pipeline} illustrates the GUIDER pipeline; Section \ref{section:Method} elaborates further on each module.

\noindent Our main contributions include:
\\ \textbf{1)} A unified probabilistic framework that continuously predicts both navigation (area-level) and, downstream from it, manipulation (object-level) intent within a coupled belief structure in the global workspace.  
\\ \textbf{2)} The Synergy Map, which combines controller input and map context to infer navigation goals without requiring predefined waypoints.
\\ \textbf{3)} Direct integration of multi-level grasp affordance (including geometric and morphological checks) fused into an object-intent layer, also without the need for predefined goals.  
\\ \textbf{4)} A quantitative evaluation across multiple human teleoperation trials, showing performance under ambiguity and task transitions.

While the current evaluation relies on post-hoc replay of live teleoperation data, GUIDER lays the groundwork for future integration into real-time teleoperation and shared-control systems. Reliable intent inference will permit transitions to different levels of autonomy and control, such as automated grasp execution or autonomous path planning.  The dual-phase design is inspired by human motor control, mirroring the spatial-to-focal reasoning observed when people perform integrated navigation and manipulation tasks \cite{cisek2007affordance}.

\section{Related Work}
Robotic intent inference draws on three main research streams, especially from the mobile manipulation perspective: probabilistic filters that estimate operator goals, constraint-aware motion and grasp planning, and human-inspired techniques. In this section, we first outline the application domains, then review each stream and how they motivate our approach for a phased framework that first narrows the likely area of action and then focuses on the specific object of interest.

\paragraph*{Application context}  
Key application domains include cooperative assembly, where robots and humans work side-by-side \cite{li2022review}; personalised assistance to individuals with limited mobility \cite{vanhooydonck2003shared, gopinath2020active};  inspection, exploration, and dismantling within hazardous environments such as those encountered in nuclear decommissioning operations \cite{santos2022robbe, marturi2016towards, wongproposed}, disaster response scenarios, and remote infrastructure maintenance \cite{panagopoulos2021bayesian, chiou2016experimental, chiou2022robot, contreras2025mini}; and flexible manufacturing operations that need adaptability to rapidly changing requirements and high precision in task execution \cite{rahman2016weight}. In these scenarios, anticipating the operator’s intentions with high confidence, namely the specific objectives the human user seeks for the robot to accomplish, is essential. 

\subsection{Navigation-intent methods}
Several methods concentrate exclusively on mobile navigation intent \cite{panagopoulos2021bayesian, best2015bayesian, yoon2021predictive}, inferring an operator’s destination or planned path by probabilistic filtering \cite{panagopoulos2021bayesian, gopinath2020active} or Hidden-Markov models (HMMs) \cite{aarno2008motion, kelley2008understanding}.  Bayesian filters that fuse present and past controller data can sharpen these estimates \cite{best2015bayesian, panagopoulos2021bayesian}, yet they typically assume a set of goals is already known. GUIDER, on the contrary, can work without priorly set goals or objectives.

\subsection{Manipulation-intent methods}
Manipulation-centered frameworks predict which object the operator intends to grasp and how to grasp it \cite{oh2021system, mainprice2013human}.  They rely on cues such as hand-gesture recognition \cite{oh2021system}, vision-based object classification \cite{wu2021spatial}, or tactile sensing \cite{shao2024constraint} to identify target objects and candidate grasps.  Learning based predictors extend this idea by forecasting human limbs motion \cite{shao2024constraint, mainprice2013human, yue2022human, ding2011human, li2022review} or future trajectories of robots and objects \cite{song2024robot, luo2022human}.  While effective within their domain, these approaches rarely bridge navigation and manipulation within a unified model, leaving a gap that the present work addresses. GUIDER fills this gap by integrating the navigation belief and the manipulation layer into a single, continuous pipeline.


\subsection{Probabilistic Intent Inference}
\paragraph*{Intent-Recognition Filters}
Robotic intent recognition is usually framed as a probabilistic time series.  
Hidden Markov Models, Dynamic Bayesian Networks (DBNs) and Inverse Reinforcement Learning (IRL) can track human actions, but they assume a fixed list of candidate goals \cite{panagopoulos2021bayesian,kelley2008understanding,aarno2008motion,tahboub2006intelligent}.  Later work relaxed that limit. Best et al. infer both destination and path with a Bayesian filter that treats the goal as a hidden variable \cite{best2015bayesian}. Yoon and Sankaranarayanan encode each possible intent as a short temporal-logic formula and keep a Bayesian belief over those formulas \cite{yoon2021predictive}.  
Most methods, however, still rely on an explicit catalogue of possibilities. In contrast, our approach maintains a dense belief over the whole map and smoothly shifts from area-level (navigation) to object-level (manipulation) intent without predefined goals, producing both \emph{where} the operator is heading and \emph{what} the object of their intent is. 

\paragraph*{Trajectory Forecasting and Dynamic Systems}
Intent inference often begins with the question \emph{where will the operator move next?} \cite{best2015bayesian}.  Early approaches used hand-crafted motion models. Multiple methods have been used for forecasting or understanding movements, applicable to both mobile and manipulation systems. For example Dynamic Movement Primitives (DMPs) represent a trajectory as the attractor of a few ordinary differential equations learned from demonstrations \cite{ijspeert2013dmp}. Template matching has been used to align current motions to stored examples with multidimensional Dynamic Time Warping (DTW) \cite{vakanski2012pbddtw}. Recent work has moved to deep learning: with conditional variational auto-encoders (CVAEs) and spatiotemporal graph networks predicting multiple future paths that a robot might take \cite{ivanovic2019trajectron, song2024robot}, and denoising diffusion models further reduce forecast error on large datasets \cite{jiang2023motiondiffuser}.  GUIDER taps into these trajectory forecasting ideas at two scales, using multi-horizon velocity projections for navigation and short-range end-effector kinematic predictions for manipulation, so trajectory forecasting informs both \textit{where} the base is heading and \emph{where} the arm is about to move.

\subsection{Constraint Handling}
At the manipulation scale, local checks usually suffice as constraints, answering the question \emph{is it possible to do the action, how?}. Constraint-aware grasp planners discard grasps that exceed the gripper aperture or joint limits \cite{shao2024constraint}, and affordance networks predict object-centred masks that highlight feasible grasp regions \cite{mandikal2021learning}.  
For the manipulation step, GUIDER adopts a local-constraint philosophy, primarily through multi-level grasp feasibility analyses (geometric, morphological, etc.), thereby lowering the probability of goals failing at any level.

\subsection{Human Factors and Shared Control}
Neuroscience suggests that people plan manipulation in two phases.  
They first choose \textit{where} or \textit{what} to act on, then refine the \textit{how} of the movement. This idea appears in the affordance competition model \cite{cisek2007affordance} and the internal-model framework \cite{wolpert1995internal}. Eye-tracking shows that gaze locks onto the future grasp target before the hand moves \cite{land2001ways,johansson2001eye}. Psychophysical studies relate object size, shape and orientation to the chosen grasp type \cite{tucker1998viewing,jeannerod1999visuomotor}. GUIDER mirrors this sequence.  Saliency masks together with the Synergy Map provide the high-level “where”, while the three-level grasp test supplies the fine-grained “how”, down-weighting objects whose geometry clashes with the gripper. The resulting intent estimates are mechanically feasible and saliency-guided.

Confidence-based shared-autonomy blends human commands with robot assistance, using different mechanisms such as Bayesian filters or set rules to decide when to help \cite{panagopoulos2021bayesian, chiou2022robot}. Although we currently evaluate with a \textit{post-hoc} replay of the teleoperation data rather than providing live assistance, dual-phase intent prediction serves as the inference backbone for future Variable Autonomy modules.

Existing studies typically cover navigation intent, manipulation intent, or isolated constraint methods, but none unify base-motion context, perceptual affordances, and fine manipulator kinematics within a single probabilistic structure that runs in real-time. GUIDER fills these gaps with its integrated design.

\section{Method} \label{section:Method}
In this section, we present GUIDER and its components for inferring human operator goals during mobile manipulation tasks. The approach processes sensor streams and inputs to incrementally refine goal probabilities through a dual-phase probabilistic process, distinguishing between navigation-focused intent estimation (interaction area) and manipulation-focused intent estimation (target object). Figure \ref{fig:intent_pipeline} summarizes the system architecture. Input data includes occupancy maps (static or dynamically updated), RGB and depth camera streams, robot odometry, and end-effector trajectories with derived velocity and acceleration.

When the operator moves the mobile base toward, say, a workbench, the system raises a probability or "hot spot" around that area, even before manipulation starts. This effect is generated during the \textbf{navigation phase}, where a probabilistic map predicts the intended interaction \textit{area}. A base layer stores free space, static obstacles, and regions around isolated objects, with values that decay exponentially over time. A motion layer then projects the current velocity of the robot forward at several short horizons. These layers combine into a synergy layer that diffuses likelihood outward from the predicted footprints into nearby regions, producing probability fields over candidate areas, and later merged with the other layers to make the Synergy Map.

When reaching the intended region, a fully autonomous \textbf{scene acquisition} sub-phase occurs. The manipulator performs an automated sequence of movements around the workspace (programmed to take roughly 35 seconds), merging point clouds into a dense local 3D model. Concurrently, a representative RGB image and raw depth map are stored. The obtained 3D scan is rendered in the operator interface to improve situational awareness before teleoperation resumes.

Picture the wrist pausing above a cluttered bench; the system must decide whether the operator intends to pick the nearby drill, marker, or box. The \textbf{manipulation phase} first uses the RGB frame to compute visual saliency with U$^{2}$-Net\cite{u2quin2020} and FastSAM\cite{fastsam2023} (prompted by clustered center points from HDBSCAN \cite{mcinnes2017accelerated}). For every salient object from this check, geometric feasibility checks based on tool compatibility are performed. A coarse gaze cue (based on the camera optical axis, boosting probability in clusters near the image center) and distance weighting are then applied to refine the probabilities further. The perceptual phase of the manipulation occurs while the automated scanning takes place, always lasting less than the hardcoded autonomous procedure. As teleoperation resumes, object probabilities evolve frame by frame using a heuristic based on gripper distance, velocity, acceleration, and small biases protecting tool-compatible objects. The outcome is a smooth, kinematics-aware belief distribution over candidate goals.

The remainder of this section first formalises the layered probabilistic map that drives the navigation phase (Section \ref{subsec:nav_phase_formulation}) area inference, then describes the autonomous scene acquisition (Section \ref{subsec:scanning_context}) scan, and finally details the perception and kinematics cues used to infer object-level intent during the manipulation phase (Sections \ref{subsec:manip_perception}, \ref{subsec:manip_eef_evolution}).

\subsection{Navigation Phase: Area Intent Representation} \label{subsec:nav_phase_formulation}

During periods with primarily mobile base motion, intent over the target interaction \textit{area} is represented using a layered 2D belief map \(B(x,y,t)\) in the global frame \(\mathcal{G}\). This map dynamically integrates information from base odometry and the occupancy grid \(\text{Occ}(x,y)\), which uses a cell size of $\Delta_{cell}$.
The base belief layer is initialized as:
\[
\small
B_{\text{base}}(x,y,0)=
\begin{cases}
0.02, & \text{Occ}(x,y)=0 \; (\text{free}),\\[2pt]
0, & \text{Occ}(x,y)=-1 \; (\text{unknown}),\\[2pt]
1, & \text{Occ}(x,y)=1 \; (\text{occupied}).%
\end{cases}
\]
The base layer \(B_{\text{base}}(x,y)\) stores this context with free-space and object-region beliefs decaying at $\gamma_{\text{base,free}}$ and $\gamma_{\text{base,obj}}$, respectively over time. Occupied cells likely to be objects are linearly inflated, giving higher belief values \([0.2,0.6]\) within radius \(r_{\text{infl}}\), around potential interaction candidates.

\begin{table*}[hb]
  \centering
  \caption{Navigation-phase hyper-parameters}
  \label{tab:nav_params}
  \begin{tabular}{@{}p{0.22\linewidth}p{0.09\linewidth}p{0.09\linewidth}%
                  p{0.26\linewidth}p{0.09\linewidth}p{0.09\linewidth}@{}}
    \toprule
    \textbf{Parameter}              & \textbf{Symbol} & \textbf{Value} &
    \textbf{Parameter}              & \textbf{Symbol} & \textbf{Value} \\ \midrule
    Cell size                       & $\Delta_{cell}$              & 0.05 m &
    Free-space decay                & $\gamma_{\text{base,free}}$ & 0.55 \\
    Object-region decay             & $\gamma_{\text{base,obj}}$  & 0.50 &
    Inflation radius                & $r_{\text{infl}}$           & 0.75 m \\
    Prediction step           & $\Delta t_{\text{pred}}$ & 0.10 s &
    Update-distance threshold & $d_{\text{upd}}$         & 0.30 m \\
    Motion radius                   & $d_{\max}$                  & 1.00 m &
    Motion evidence threshold       & $\beta$                     & 0.01 \\
    Motion-layer decay              & $\gamma_{\text{decay}}$     & 0.25 &
    Motion blend factor             & $\lambda$                   & 0.40 \\
    Synergy radius                  & $r_{\text{syn}}$            & 0.15 m &
    Synergy-layer decay             & $\gamma_{\text{syn}}$       & 0.75 \\
    Synergy lower bound             & $\eta_{\text{lb}}$          & 0.60 &
    Synergy reset value             & $\eta_{0}$                  & 0.70 \\
    Synergy increment               & $\eta_{\text{inc}}$         & 0.05 &
    Synergy cap                     & $\eta_{\text{cap}}$         & 0.90 \\
    \bottomrule
  \end{tabular}
\end{table*}

A motion evidence layer, \(B_{\text{motion}}(x,y,t)\), captures the navigational commands given by the operator. Given the base odometry \((x,y,v_x,v_y)\), we linearly extrapolate its planar velocity to predict future poses \((x_p,y_p)\) at three horizons \(\tau_i\in\{5,10,30\}\,\text{s}\). Trajectories are resampled every \(\Delta t_{\text{pred}}\) and layers refresh only after the base moves \(d_{\text{upd}}\) to manage computational resources, ensuring updates correspond to significant steps, above the usual odometry jitter (typically < 0.15m). For each pose, radial evidence
\begin{equation}\label{eq:radial-evidence}
E_{\tau_i}(x,y)=\max\!\Bigl(0,1-\tfrac{\lVert(x,y)-(x_p,y_p)\rVert_2}{d_{\max}}\Bigr)
\end{equation}
is assigned within radius \(d_{\max}\), decreasing linearly from 1 at the predicted point to 0 at \(d_{\max}\). The multi-horizon blend
\begin{equation}\label{eq:multi-horizon-blend}
\tilde E(x,y)=\max_i\{\alpha_i\,E_{\tau_i}(x,y)\},
\end{equation}
with weights \(\alpha_i=\{0.60,0.60,0.85\}\), which favours the farthest projection during steady driving.  Between updates, the layer decays by the factor \(\gamma_{\text{decay}}\) to prioritize recent commands; at each update we clip
\[ B_{\text{motion}}\leftarrow\operatorname{clip}\bigl(B_{\text{motion}}+\lambda\tilde E,0,1\bigr),
\quad\lambda=0.40.
\]
The \textit{synergy} layer \(B_{\text{syn}}(x,y,t)\) highlights regions where predicted motion aligns with potential interaction sites. Seed cells satisfy \(B_{\text{base}}\ge0.2\) and \(B_{\text{motion}}\ge\beta\). A flood-fill breadth‐first search (BFS) spreads belief from each seed up to radius \(r_{\text{syn}}\), mirroring how operators steer toward a landmark while correcting drift. When visiting a cell, we apply
\[
B_{\text{syn}}^{t} =
\begin{cases}
\eta_{0}, & \text{if } B_{\text{syn}}^{t^-}\!<\eta_{\text{lb}},\\[2pt]
\min\!\bigl(B_{\text{syn}}^{t^-}\!+\eta_{\text{inc}},\eta_{\text{cap}}\bigr),
& \text{otherwise},
\end{cases}
\]

and decay the layer between odometry updates by \(\gamma_{\text{syn}}\). Where \(t^{-}\) denotes the value immediately before the update. The resulting \textit{area} belief, called the Synergy Map or combined base belief, is the point-wise maximum of the base, motion, and synergy layers:
\begin{equation}\label{eq:combined-belief}
\scriptsize
B_{\text{combined}}(x,y,t)=\max\{B_{\text{base}},\,B_{\text{motion}},\,B_{\text{syn}}\}
\end{equation}

\begin{figure}[h]
    \centering
    \includegraphics[width=1.00\linewidth]{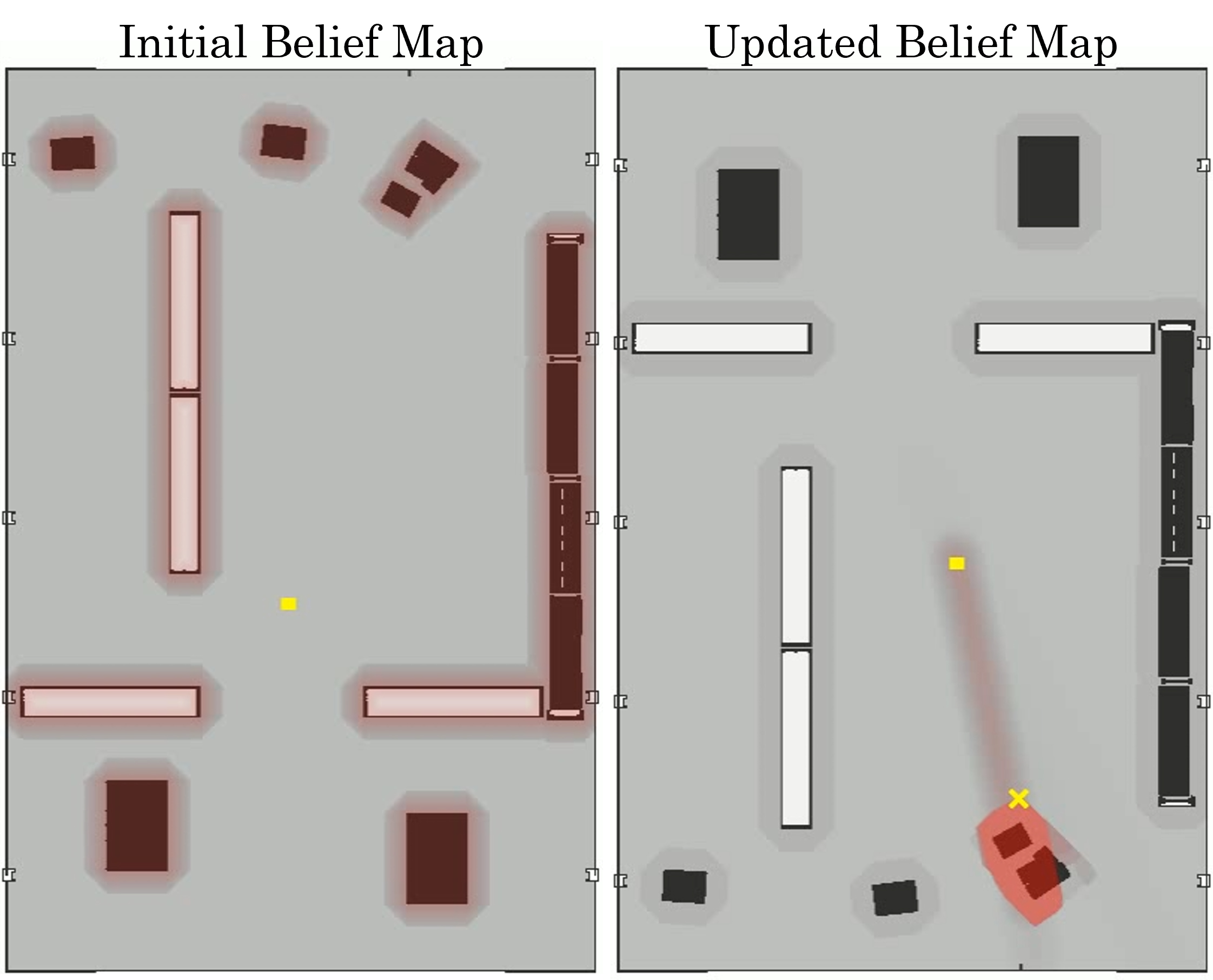}
    \caption{ Left: initial base belief ($B_{\text{base}}$) with higher values (red) in inflated object zones. Right: The combined belief map after incorporating motion and synergy updates, showing a focused "hot spot" with the peak likelihood (yellow "X") that indicates the predicted goal. The yellow dot indicates the robot's position at the time of the predictions.}
    \label{fig:base_motion_synergy}
\end{figure}

\noindent Hyperparameters were manually tuned, and are fixed as in Table \ref{tab:nav_params}. Figure \ref{fig:base_motion_synergy} shows a comparison of the belief map at the start of the task and the belief map after several rounds of updates.

\subsection{Scene Acquisition} \label{subsec:scanning_context} 
\begin{figure}[h]
  \centering
  \includegraphics[width=1.0\linewidth]{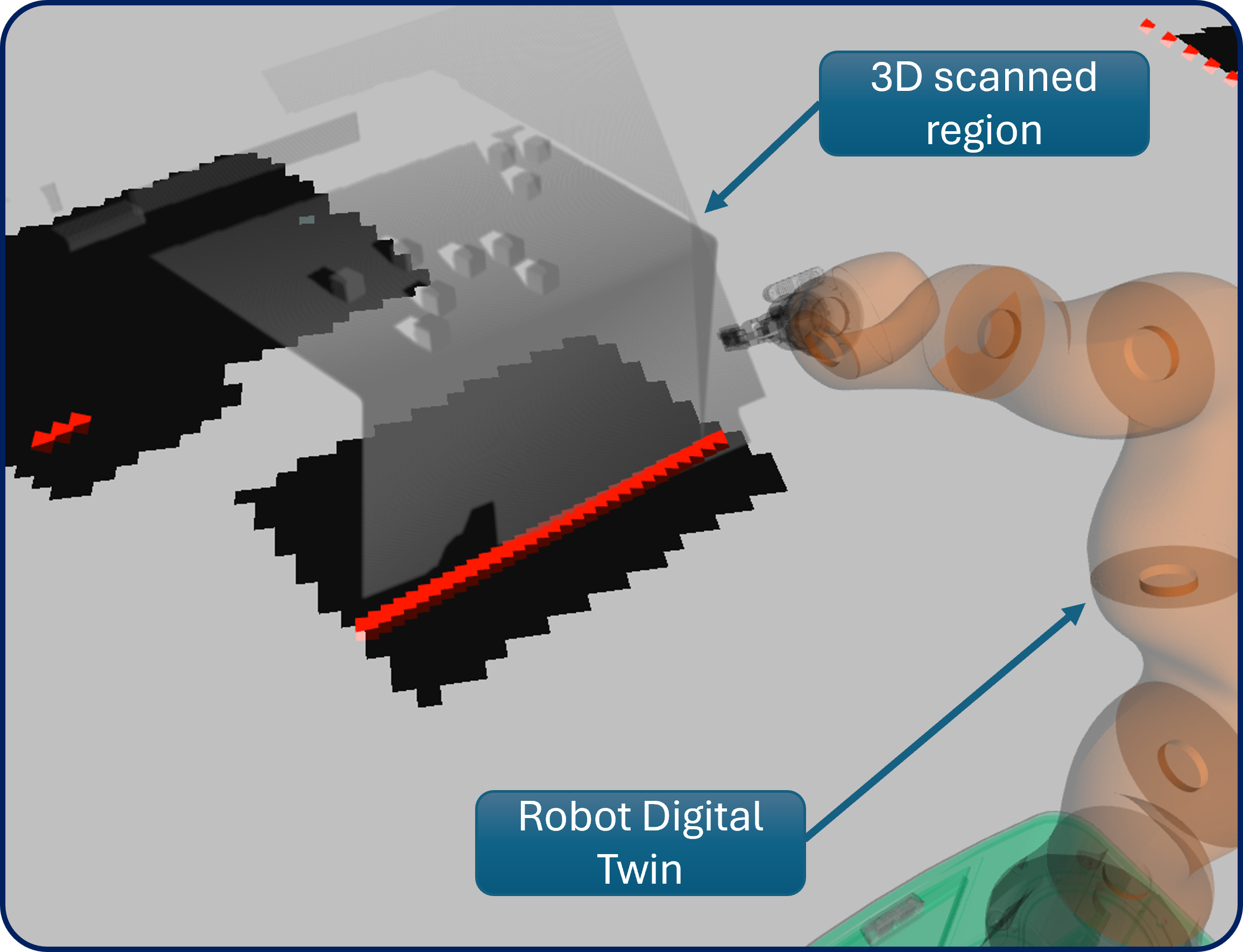}
  \caption{Fused 3D scan rendered alongside the robot digital twin to enhance operator depth perception.}
  \label{fig:three-d-reconstruction}
\end{figure}

To supply dense geometry for the manipulation phase, the operator can press a controller button that launches an autonomous six-view scan lasting \(\approx 35\) s (hardcoded) on our platform. At each viewpoint, the camera acquires a raw depth cloud \(\mathcal{P}_i\). Points closer than \(d_{\min}=0.30\) m are discarded to remove the gripper; the remainder is cropped to a workspace \(\mathcal{B}\) and expressed in the base frame by the rigid transform \(T_i\!\in\!SE(3)\) (camera-to-base). We denote the filtered cloud at pose \(i\) by \(\hat{\mathcal{P}}_i\) and the final fused model by \(\mathcal{P}_{\text{merged}}\)

\[
\hat{\mathcal{P}}_i =\{\,T_i p\mid p\!\in\!\mathcal{P}_i,\, \lVert p\rVert>d_{\min},\, p\!\in\!\mathcal{B}\},
\]
\[
\mathcal{P}_{\text{merged}} =\operatorname{Voxel}\!\Bigl(\textstyle\bigcup_{i=0}^{5}\hat{\mathcal{P}}_i,\;\ell_v=0.01\text{ m}\Bigr),
\]

\noindent where voxelisation (A leaf size \(\ell_v\)) of 1 cm suppresses sensor noise. The first RGB frame and raw-depth image are stored for use in the manipulation phase, and the fused cloud is streamed live to the operator\footnote{Early pilot testing of the robot controllers without the 3D view yielded \(3/6\) task completions; adding the fused cloud raised performance to \(6/6\) as users reported greatly improved spatial awareness in the pilot tests.} (Fig. \ref{fig:three-d-reconstruction}).

\subsection{Manipulation Phase: Perceptual Processing and Object Identification} \label{subsec:manip_perception}
During the Manipulation Phase, a perception pipeline extracts object information within the area of intent and assigns probabilistic values using heuristic features.

\subsubsection{Point Cloud representation}
\label{subsec:pc_representation}
Once detailed scene data is available and the operator's focus shifts to manipulation, starting from a single RGB-D snapshot \((I,D)\), the depth pixels are converted into a 3D point set expressed in the robot's base frame.

\paragraph{Depth reprojection} The camera’s intrinsic parameters \((f_x,f_y,c_x,c_y)\) are read directly from the accompanying \texttt{/camera\_info} message.  For each pixel \((u,v)\) with valid depth \(z=D(u,v)\), the pinhole model gives

\[
X=\Bigl(\tfrac{u-c_x}{f_x}\Bigr)z,\qquad
Y=\Bigl(\tfrac{v-c_y}{f_y}\Bigr)z,\qquad
Z=z,
\]

yielding a set of points \(\{\mathbf{p}_i^{\mathrm{cam}}\}\subset\mathbb{R}^3\).

\paragraph{Point transformation and filtering}
Let \(\mathbf{T}_{F_{\mathrm{src}}\!\to F_{\mathrm{tgt}}}\in\mathrm{SE}(3)\) be the timestamped TF transform. Each homogeneous point \(\tilde{\mathbf{p}}=[X,Y,Z,1]^\top\) is mapped to the target frame:

\[
\tilde{\mathbf{p}}^{\mathrm{tgt}}
     =\mathbf{T}_{F_{\mathrm{src}}\!\to F_{\mathrm{tgt}}}\;
      \tilde{\mathbf{p}}^{\mathrm{cam}}.
\]

Points outside the application's depth band \(Z\!\notin\![Z_{\min}, Z_{\max}]\) are discarded, removing gripper artifacts and distant clutter. The resulting cloud forms the geometric input for subsequent stages.

\subsubsection{Saliency with U$^{2}$-Net}
\begin{figure}[h]
    \centering
    \includegraphics[width=1.00\linewidth]{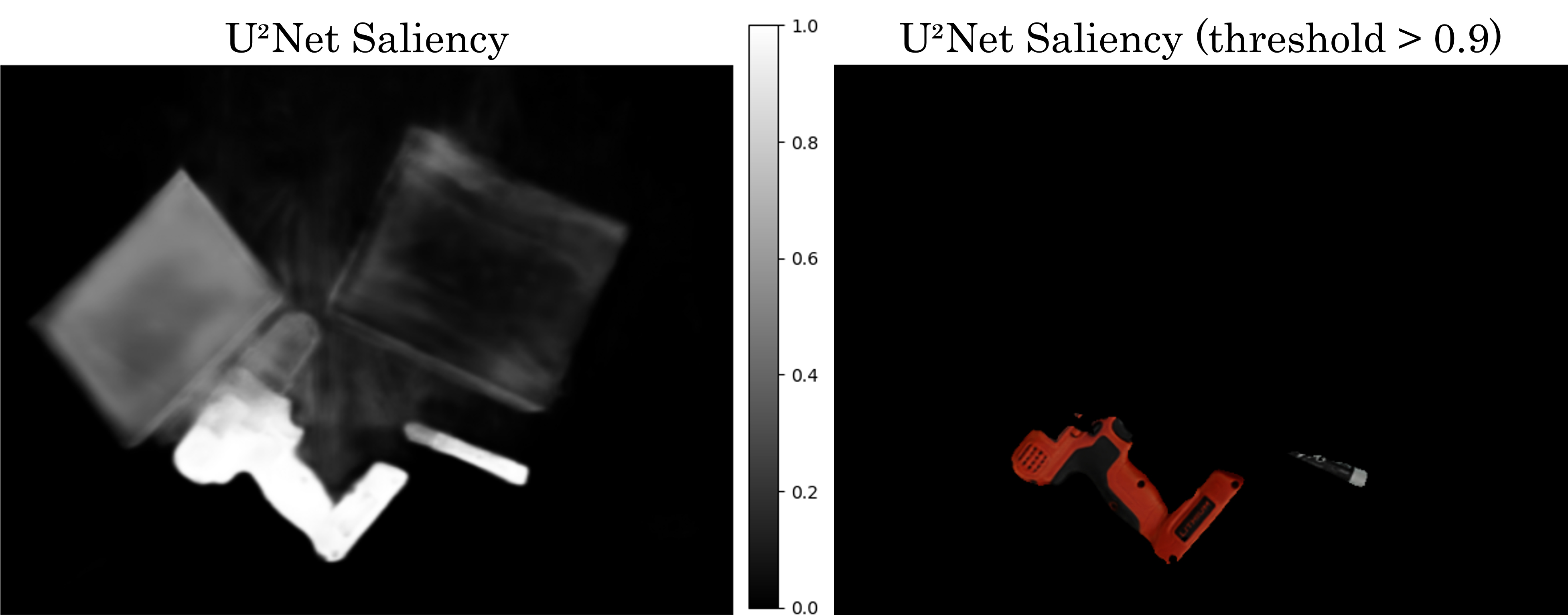}
    \caption{U$^{2}$-Net output. Left: greyscale saliency map $\mathbf{S}$. Right: original RGB masked by binary mask $\mathbf{B}$ ($\tau=0.9$).}
    \label{fig:u2net_saliency}
\end{figure}

Visual saliency marks the parts of an image that naturally draw a human’s gaze and therefore make good first guesses about what the operator may want.  We pass the RGB frame $\mathbf{I}\!\in\!\mathbb{R}^{H\times W\times3}$ through a pre-trained U$^{2}$-Net, chosen due to its capacity to capture edges and saliency in a single pass.  The image is resized to $320\times320$, normalized with standard ImageNet values.  The network returns a single-channel map $\mathbf{S}\!\in\![0,1]^{H\times W}$; we stretch it to the full 0-1 range by

\[
\mathbf{S}(u,v)\leftarrow\frac{\mathbf{S}(u,v)-\min(\mathbf{S})}{\max(\mathbf{S})-\min(\mathbf{S})+\varepsilon},\quad\varepsilon=10^{-8},
\]

\noindent then apply an empirically chosen threshold $\tau=0.9$ (chosen to retain only high-confidence regions) to form a binary mask $\mathbf{B}$ with $\mathbf{B}(u,v)=1$ when $\mathbf{S}(u,v)>\tau$.  The continuous map $\mathbf{S}$ is used as an initial understanding of the objects in the scene, while $\mathbf{B}$ is used later in the probability fusion step. Figure \ref{fig:u2net_saliency} shows the original U$^{2}$-Net saliency mask, and the thresholded mask overlayed on the original RGB image.

\subsubsection{Plane Removal and Clustering}
To generate 3D points of interest for prompting a subsequent instance segmentation network FastSAM, we first process the point cloud derived from the depth image. This involves removing the dominant planar surface and then clustering the remaining points to identify potential object locations.

\paragraph{Plane segmentation} Given a cloud $\{\mathbf{x}_i\!\in\!\mathbb{R}^3\}$ we down-sample with a $2\,$mm voxel grid and fit a single plane \\
\[a x + b y + c z + d = 0\] 
by RANSAC (distance threshold $8.5$ mm, $r\!=\!16$ samples, $2\,000$ iterations). Points whose signed distance to that plane is $\le\!5$ mm are discarded, leaving the non-planar set \(\mathcal{P}_{\mathrm{obj}}\). These numbers match the simulated RealSense depth noise at 1 m. If the plane fit misses a small area, the subsequent saliency fusion, and ultimately the motions given by the human operator, can still compensate for the errors.

\begin{figure}[htbp]
  \centering
  \includegraphics[width=0.99\linewidth]{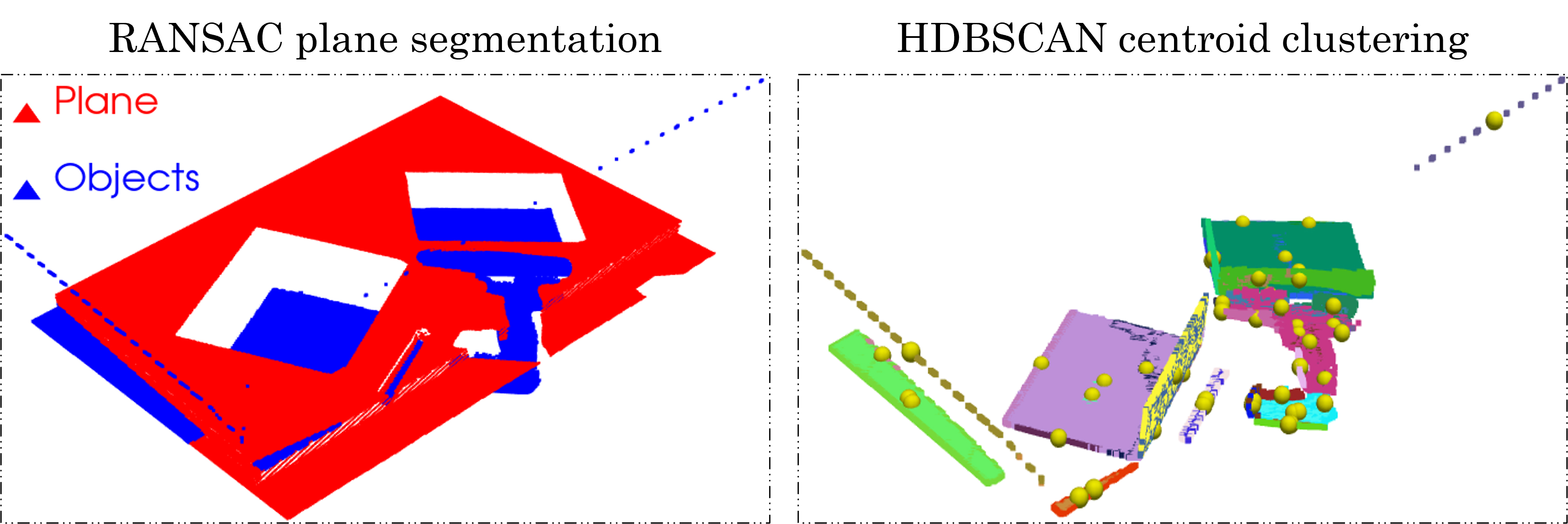}
  \caption{Key stages in object identification. Left: Ground plane and object points identified via RANSAC plane segmentation. Right: Object clusters and centroids generated by HDBSCAN.}
  \label{fig:clustered_planar_results}
  \vspace{-0.4\baselineskip}
\end{figure}

\paragraph{HDBSCAN clustering \cite{mcinnes2017accelerated}} For each point $\mathbf{p}_i$ we compute a surface normal \(\mathbf{n}_i\) (search radius $5$ mm, $k_{\mathrm{nn}}=5$) and build 
\[
\mathbf{f}_i=\bigl[x_i,\,y_i,\,z_i,\,\alpha n_{x},\,\alpha n_{y},\,\alpha n_{z}\bigr],
\qquad\alpha=0.1,
\] 

so orientation differences influence the distance metric without dominating the Euclidean term. HDBSCAN with \texttt{min\_cluster\_size}$=15$ then produces a set of clusters $\{\mathcal{C}_k\}$ and centroids \(\mathbf{c}_k=\frac{1}{|\mathcal{C}_k|}\sum_{\mathbf{p}\in\mathcal{C}_k}\mathbf{p}\) that later serves as image-plane prompts for FastSAM. 
Figure \ref{fig:clustered_planar_results} shows an example of the plane segmentation and clustering steps.

\subsubsection{Saliency from FastSAM}
Even after 3D clustering, many local point groups may belong to the same physical object, just as humans first glimpse separate edges or corners and then recognise a single whole. To reproduce that behaviour we project \emph{every} cluster centroid \(\mathbf{c}=(X,Y,Z)\) to image coordinates

\[
u=\frac{f_x X}{Z}+c_x,\qquad
v=\frac{f_y Y}{Z}+c_y,
\]

and feed the full set of projected points as foreground prompts to a single FastSAM forward pass. Because a centroid almost always lies inside its object, multiple prompts landing on the same item naturally collapse into one continuous mask.

FastSAM returns one mask per prompt. We keep all masks except those covering more than 25\% of the image, which are treated as segmentation errors. Masks whose FastSAM confidence falls below a fixed threshold ($\tau_{\text{conf}}{=}0.4$) are also rejected; this removes prompts on background clutter or image borders that FastSAM itself deems unreliable. Let $M_i^{\star}$ denote these remaining valid masks. They are merged via the logical function OR to obtain the final binary saliency mask $\mathbf{F}$:

\[
\mathbf{F}(u,v)=\bigvee_i M_i^{\star}(u,v).
\]

\begin{figure}[htbp]
  \centering
  \includegraphics[width=1.00\linewidth]{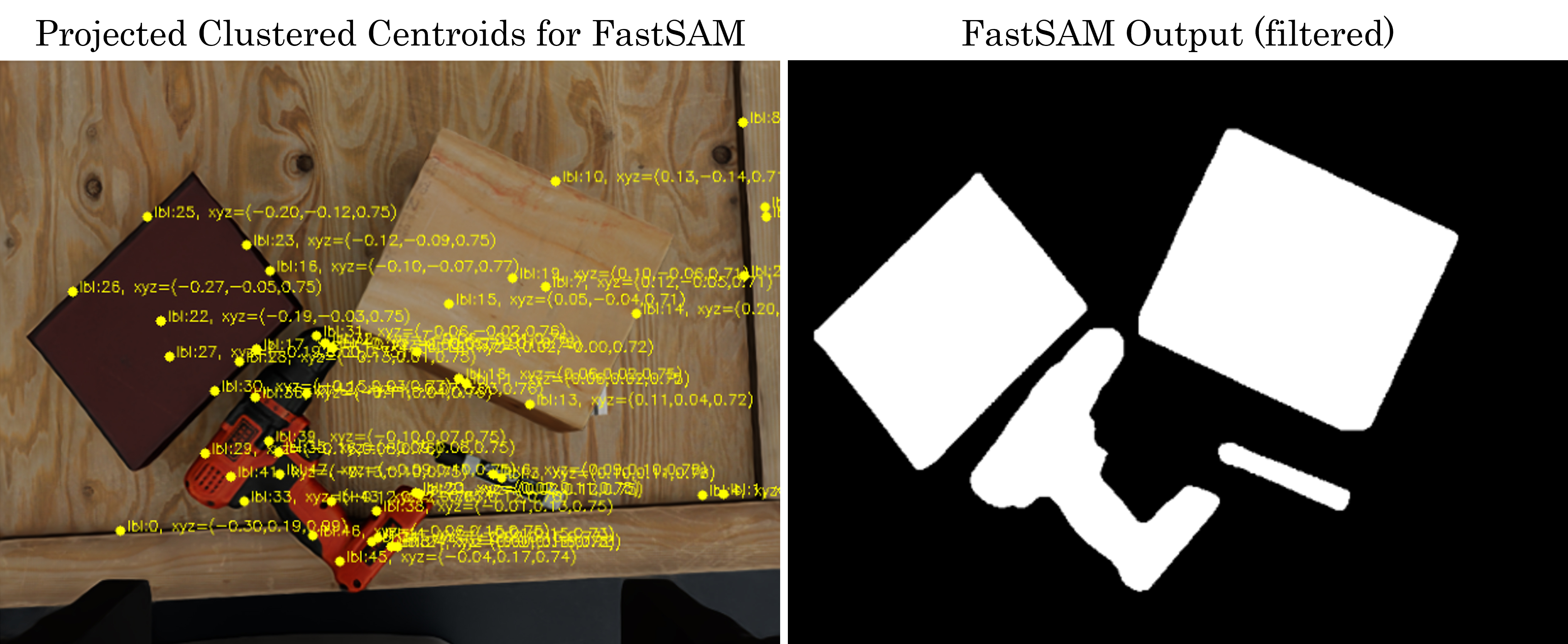}
  \caption{FastSAM segmentation using projected 3D cluster centroids. Left: Input prompts overlaid on the RGB image. Right: Output of consolidated object masks after segmentation.}
  \label{fig:clustered_points_fastsam}
\end{figure}

This two-step strategy preserves every plausible sub-feature but quickly reduces them to a smaller set of coherent objects that drive the subsequent probability-fusion stage. Figure \ref{fig:clustered_points_fastsam} shows the FastSAM step, with the RGB image and cluster prompts on the left and the merged object masks on the right.

\subsubsection{2D Saliency Fusion} \label{sec:merged_saliency}
Just as two independent visual cues make a region stand out more than either cue alone, we fuse the binary FastSAM and U$^{2}$-Net masks to weight pixels by how many detectors agree. 

For each image coordinate \((u,v)\) we assign a saliency weight \(P(u,v)=0\) if neither mask is active, \(P(u,v)=0.6\) when exactly one mask is active, and \(P(u,v)=0.9\) when both agree. Because the detectors use different backbones (transformer vs.\ encoder-decoder), their agreement provides a stronger cue than either detector alone.  A depth buffer (z-buffer) for every salient pixel is stored alongside \(P\) to enable later 3D consistency checks.

\subsubsection{Logical Constraints and Tool Affordances} \label{sec:logical_constraints}
We first test whether each segmented object can be pinched by a parallel-jaw gripper whose maximum opening is \(2r = 85\text{ mm}\) (\(r = 0.0425\text{ m}\)).  Changing \(r\) adapts the rule to any other tool width.

\paragraph{Bounding-box test}
For every object mask we extract the oriented 2D bounding box \((w_{\text{px}}, h_{\text{px}})\) with \textsc{minAreaRect}.  Using the object’s median depth \(Z_{\text{obj}}\) and the focal length \(f_x\) we estimate a metric scale

\[
\gamma = \frac{Z_{\text{obj}}}{f_x}, \qquad
w_{\text{m}} = w_{\text{px}}\gamma,\;
h_{\text{m}} = h_{\text{px}}\gamma .
\]
An object is marked \emph{bounding-box feasible} for our tool when \(\min(w_{\text{m}}, h_{\text{m}}) \le 2r .\) reflecting whether the short side fits within the parallel jaws’ maximum opening, and the mask of the feasible object is stored in $\mathbf{M}_{\text{bbox}}$. This 2D check is conservative, and further checks can provide more accurate estimates.

\paragraph{Morphological approximation of tool geometry} Although our end effector is a parallel-jaw gripper, we bound its \(85\,\mathrm{mm}\) jaw opening by a circle of radius \(r = 42.5\,\mathrm{mm}\).  For an object mask \(\mathbf{M}\in\{0,1\}^{H\times W}\) we estimate a depth-dependent scale \(\gamma = Z_{\text{obj}}/f_x\).  Let \(\mathbf{K}_{r/\gamma}\) denote a \textbf{binary disk structuring element} whose radius is \(r/\gamma\) pixels.
A single erosion
\[
\mathbf{M}_{\text{eroded}}
   = \mathbf{M} \ominus \mathbf{K}_{r/\gamma}
\]
suffices: if \(\max(\mathbf{M}_{\text{eroded}})=0\), the object is fully eliminated, then the object fits between the jaws and is flagged \emph{morph-feasible} and the mask is stored in $\mathbf{M}_{\text{morph}}$. Because a circle can circumscribe any planar tool footprint, the same test can be applied to other tools or 2-finger grippers by changing \(r\).

\paragraph{Advanced geometry check}
As a third, more detailed check, to ensure the virtual jaws fit inside locations within the silhouette of an object, we generate a finite candidate set

\[
\mathcal{S}=
\underbrace{\mathrm{V}\!\bigl(\mathrm{Hull}(\mathcal{C}_0)\bigr)}_{\text{convex-hull vertices}}
\;\cup\;
\bigcup_{i=1}^{N_\ell}
\mathrm{V}\!\bigl(\mathrm{Hull}(\mathcal{L}_i)\bigr)
\;\cup\;
\mathcal{O},
\]

where  
\(\mathcal{C}_0\) is the contour of the full mask, \(\mathcal{L}_i\) are the contours of the leftover components produced by circular erosion, \(\mathrm{Hull}(\cdot)\) is the 2D convex hull, \(\mathrm{V}(\cdot)\) outputs its vertices, and \(\mathcal{O}\) collects \textit{opposite corners} obtained by raycasting from each hull vertex through the centroid and recording the first contour hit on the far side.

Duplicate points are removed by rounding to two decimals. Unordered pairs whose midpoints differ by less than \(d_{\text{skip}}=1\) are merged.

\begin{figure}[!h]
    \centering
    \includegraphics[width=1.00\linewidth]{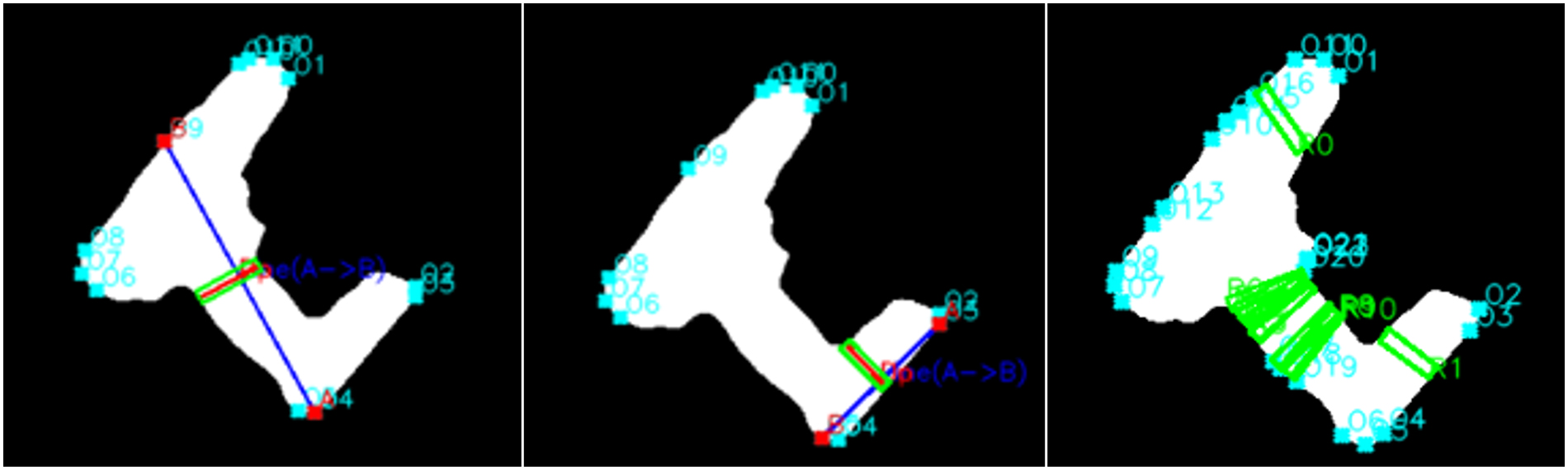}
    \caption{Visualization of the advanced geometry check for grasp affordance. The algorithm generates candidate grasp pairs across object contours and validates them, showing the final set of feasible grasp rectangles on a representative object.}
    \label{fig:advanced_representation}
\end{figure}

(1) For every remaining pair \((\mathbf{p}_A,\mathbf{p}_B)\) we form the midpoint \(\mathbf{M}\) and inward normal
      \[
      \mathbf{n}=(-(y_B-y_A),\,x_B-x_A)
      \]
Running from \(\mathbf{M}\) along \(\pm\mathbf{n}\) gives the normal clearance \(\Delta_{\!\perp}\) (px). With pixel-to-metre scale \(\gamma=Z_{\text{obj}}/f_x\) the pair is discarded if \(\Delta_{\!\perp}>2r/\gamma\).
(2) If the opening fits, we place (a) a main rectangle of length \(\Delta_{\!\perp}\) and height of finger thickness \(w_f=8.25\;\mathrm{mm}\), (b) a clearance rectangle of length \(\Delta_{\!\perp} + 2\,\mathrm{mm}\) that ensures free space for the last millimetres of jaw travel expressed in pixels via \(\gamma\). Let \(\mathrm{Cov}_{\mathrm{main}}\) be the fraction of the main rectangle covering the object, and \(\mathrm{Cov}_{\mathrm{extra}}\) be that of the extended rectangle. We declare the grasp feasible if 
      \[
      \mathrm{Cov}_{\text{main}}\ge0.99,
      \qquad
      \mathrm{Cov}_{\text{extra}}<0.95.
      \]
Generated rectangles and contour pairs on sample data are shown in Figure \ref{fig:advanced_representation} for a visual representation of the expressed requirements. This specific check generates two masks: one for the object level, $\mathbf{M}_{\text{adv-obj}} $, and one for the rectangle level, $\mathbf{M}_{\text{adv-rect}}$, specifically for the gripping positions. The test is computed in parallel across all pairs.

\begin{figure}[htbp]
  \centering
  \includegraphics[width=\linewidth]{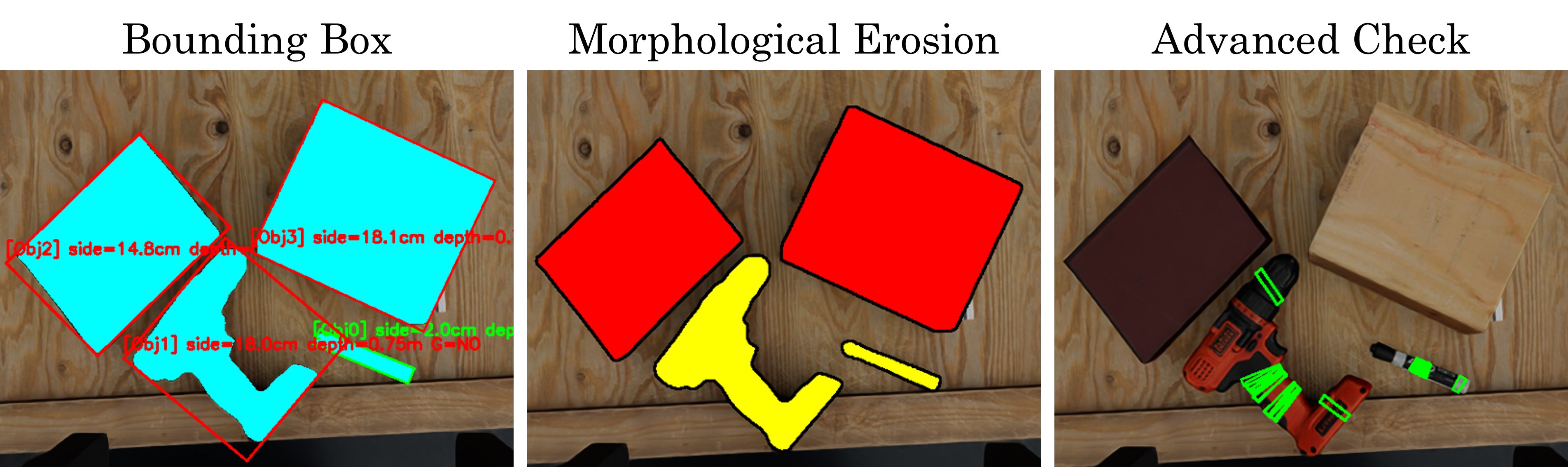}
  \caption{\textbf{Left}: Bounding Box (red = unfeasible, green = feasible) \textbf{Center}: Morphological erosion (yellow = graspable, red = too large).  \textbf{Right}: Feasible pinch rectangles (green = graspable locations) found by the advanced geometry check.}
  \label{fig:fused_geometric}
\end{figure}

The different masks are then fused down the pipeline to determine how likely it is for an operator to intend to interact with the objects using the current tool at their disposal. Figure \ref{fig:fused_geometric} shows an example output of the three feasibility tests. 

\subsubsection{Object-level saliency fusion} \label{sec:object_saliency_fusion}
The mask accumulated so far, including visual saliency and geometric feasibility, in addition to other functions such as distance and camera-centric priors, are combined into a single probability field and then compressed to one score per object.

\paragraph{Pixel level cascade.}
Let \(\mathbf{S}\in[0,1]^{H\times W}\) be the fused saliency map from Sec. \ref{sec:merged_saliency}.  Starting with \(\mathbf{P}^{(0)}\equiv\mathbf{S}\) we apply the following transformations:

\begin{figure*}[h]
    \centering
    \includegraphics[width=1.00\linewidth]{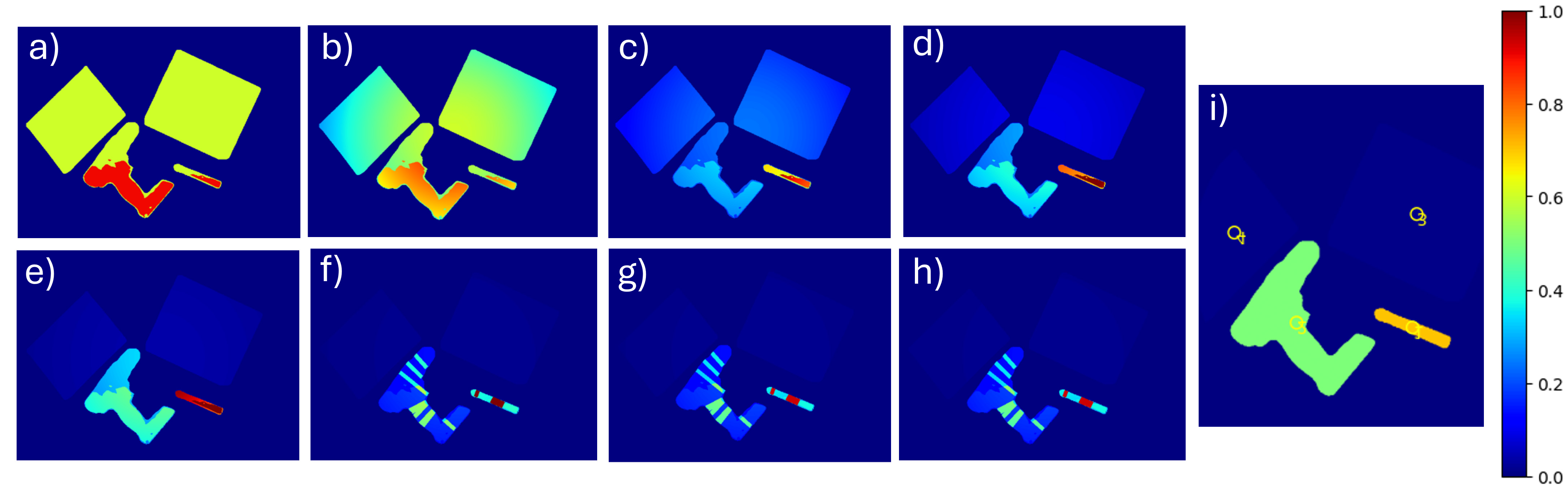}
    \caption{Intermediate probability masks after calculations:  a) Visual saliency, b) Centre weight, c) Bounding box Mask; d) Morphological mask. e) Advanced Geometry check. f) Advanced Geometry Gripping. g) Depth Weighting. h)  Flooring. i) Final Rescaled Mask.}
    \label{fig:masked_outputs}
\end{figure*}

\noindent (i) \textbf{\textit{Centre bias.}} A Gaussian weight  
   \[
     w_{\mathrm{c}}(u,v)=
       \exp\!\bigl[-\tfrac{(u-u_0)^2+(v-v_0)^2}{2\sigma_{\mathrm{c}}^2}\bigr]
   \]
here \((u_0,v_0)\) denotes the camera’s principal point, the image centre, with \(\sigma_{\mathrm{c}}\!=\!240\) px (matching the vertical radius of the image) down-weights peripheral regions, reflecting the human tendency to center targets in the center of the camera view. \(\mathbf{P}^{(1)} = w_{\mathrm{c}}\!\odot \mathbf{P}^{(0)}\).
    
\noindent (ii) \textbf{\textit{Mask multiplication.}} For each binary mask
   \(\mathbf{M}\in\{\mathbf{M}_{\text{bbox}}, \mathbf{M}_{\text{morph}}, \mathbf{M}_{\text{adv-obj}}, \mathbf{M}_{\text{adv-rect}}\}\) from Sec. \ref{sec:logical_constraints}.
   we scale
   \[
     \mathbf{P}^{(k+1)}(u,v)=
     \begin{cases}
       \lambda_{\text{in}}\;\mathbf{P}^{(k)}(u,v), & \mathbf{M}(u,v)=1,\\[4pt]
       \lambda_{\text{out}}\;\mathbf{P}^{(k)}(u,v), & \mathbf{M}(u,v)=0,
     \end{cases}
   \]
   with \(\lambda_{\text{in}}=1.2\) and \(\lambda_{\text{out}}=0.4\). Pixels consistent with a feasible-geometry mask are boosted; pixels that don't appear on a mask are de-emphasised but not nulled, preventing a single cue from erasing evidence supplied by others.

\noindent (iii) \textbf{\textit{Depth weighting.}} Depth \(z(u,v)\) is converted to a weight  
   \[
     w_{\!\text{d}}(u,v)=
       \exp\!\bigl[-\alpha_d\,(z(u,v)-z_0)^2\bigr],
     \qquad
   \]
   with \(\alpha_d=1.5,\;\) and the nominal working distance \(z_0=1\,\text{m},\)
   and applied only where \(\mathbf{P}^{(k)}\!>\!0.3\): \(\mathbf{P}^{(k+1)} = \mathbf{P}^{(k)}\odot w_{\!\text{d}}\). This favors objects inside the nominal working distance and leaves low-confidence pixels unchanged.

\noindent (iv) \textbf{\textit{Flooring.}} A small floor \(\varepsilon=0.01\) is imposed so that locations with any positive evidence remain visible to later temporal fusion:
   \[
     \mathbf{P}^{\ast}(u,v)=
     \max\bigl(\mathbf{P}^{(k)}(u,v),\,\varepsilon\bigr).
   \]

 \paragraph{Connected-object pooling} Pixels with \(\mathbf{P}^{\ast}\ge\tau_\ell\) (\(\tau_\ell=0.01\)) form a binary foreground; 8-connected components smaller than \(n_{\min}=10\) px are discarded to suppress speckles and a blob specific Otsu pass removes any low-probability halos that could merge adjacent objects. Each remaining component \(\mathcal{O}_i\) receives a single score

\[
  g_i = \min_{(u,v)\in\mathcal{O}_i} \mathbf{P}^{\ast}(u,v)
\]

Components are sorted by \(g_i\) and their scores are linearly rescaled to \([g_{\min},g_{\max}]\!=\![0.5,0.7]\) once they exceed a task-defined relevance threshold (\(\tau_{\mathrm{rs}}\!=\!0.05\) by default).  The result is a mask \(\mathbf{G}\) in which every object’s pixel probability is set to \(g_i\).

\paragraph{Human rationale} A centrally placed, geometry-feasible, correctly distanced object is what an operator naturally expects to grasp first; the cascade mimics this intuition through successive boosts. Soft multiplicative factors \((\lambda_{\text{in}},\lambda_{\text{out}})\) discourage but do not erase dissenting evidence, echoing a human’s willingness to reconsider alternatives. Object-level pooling collapses thousands of pixels into a handful of ranked proposals, matching how an operator thinks in terms of “which \emph{object} shall I pick?” rather than “which pixel?". Figure \ref{fig:masked_outputs} shows the output of probability masks after each step in the perceptual phase of manipulation.

\subsection{Manipulation Phase: Final Refinement via End-effector Motion} \label{subsec:manip_eef_evolution}

Let \(\{O_i\}_{i=1}^{N}\) be the connected components obtained in Sec. \ref{sec:object_saliency_fusion}, each represented by its centroid \(\mathbf{c}_i=[x_i,\,y_i,\,z_i]\in\mathbb{R}^3\) and an initial probability \(p_i(0)=g_i\in(0,1]\). During the teleoperation interval we update \(p_i(t)\) at every odometry sample to reflect the evolving likelihood that the operator intends to grasp \(O_i\).

\paragraph{Gripper workspace.} The central operation area of two-finger gripper is approximated by an upright cylinder of radius \(r_g= 0.028\,\mathrm{m}\;(=0.085/3)\) and height \(h_g= 0.10 \,\mathrm{m}\). With the tool-centre point (TCP) at \(\mathbf{q}(t)=[x_e,\,y_e,\,z_e]^{\!\top}\) the radial clearance is

\begin{equation}
\begin{split}
d_i(t) = \Bigl[ & \max(0,\rho_i(t)-r_g)^2 \\
               & + \max\bigl(0,z_{\min}-z_i,\,z_i-z_{\max}\bigr)^2 \Bigr]^{1/2}
\end{split}
\label{eq:cyl_dist}
\end{equation}

where \(\rho_i(t)=\bigl[(x_i-x_e)^2+(y_i-y_e)^2\bigr]^{1/2}\),

\(z_{\min}=z_e-h_g\), \(z_{\max}=z_e\).

\paragraph{Approach indicator.} \(\Delta d_i(t)=d_i(t)-d_i(t-\!\Delta t)\) is negative while the TCP moves towards the object \(O_i\).
The boolean of approaching \(\mathrm{app}_i(t)=\bigl[\Delta d_i(t)<0\bigr]\vee\bigl[d_i(t)<5\,\mathrm{mm}\bigr]\) simplifies the growth/decay logic.

\paragraph{Growth and decay coefficients.}
Experimentally tuned constants follow directly from the implementation: \(
\alpha_g=0.08,\;
\alpha_d=0.80,\;
\gamma_v=0.10,\;
\gamma_a=0.05,\;
\delta=0.10\,\mathrm{m}.
\)
Objects that began with minimal evidence \(p_i(0)<0.05\) receive a reduced growth coefficient \(\alpha_g/\kappa\) with \(\kappa=10\) and are capped at a maximum probability of \(p_{\mathrm{cap}}=0.30\).

\paragraph{Continuous-time model.} With speed \(v=\|\dot{\mathbf{q}}\|\) and acceleration \(a=\|\ddot{\mathbf{q}}\|\) (estimated using finite differences from the end-effector odometry sequence \(\mathbf{q}(t)\)) we define 

\begin{align}
G_i &= \frac{\alpha_g}{d_i+\delta}
      \Bigl[1+\gamma_v v+\gamma_a a\Bigr]
      \bigl(1-0.7\,\bar{\mathrm{app}}_i\bigr),\\
D_i &= \alpha_d\,(d_i+\delta)
      \Bigl[1+0.3\gamma_v v+0.3\gamma_a a\Bigr]
      \bigl(1+0.5\,\bar{\mathrm{app}}_i\bigr),
\end{align}

where \(\bar{\mathrm{app}}_i=1-\mathrm{app}_i\). The first term encourages objects that are close and being approached, whereas the second penalises distant or retreating ones.

\paragraph{Discrete update.} 
At every timestep (\(\Delta t = 0.004\,\text{s}\)) we advance each object’s belief with a single forward-Euler step:  
\begin{align}
\tilde p_i &= p_i(t-\!\Delta t) + \Delta t\bigl(G_i-D_i\bigr), \\[2pt]
p_i(t)    &= \min\!\Bigl( 
                 \max\!\bigl[\tilde p_i,\; p_i(t-\!\Delta t)-\beta_i\bigr],\;
                 p_{\max,I} 
             \Bigr). \label{eq:update_rule}
\end{align}

The first line yields a candidate probability \(\tilde p_i\) by adding the net growth \(G_i-D_i\) over the timestep. The second line clips that candidate so that it (i) cannot drop faster than the bias \(\beta_i\), and (ii) never exceeds its ceiling \(p_{\max,i}\). We set \(p_{\max,i}=0.99\) for regular objects and \(p_{\max,i}=p_{\mathrm{cap}}\) for those that started with very low evidence \(\bigl(p_i(0)<0.05\bigr)\). The bias term \(\beta_i\) prevents abrupt loss of the operator’s current focus:
\[
\beta_i=
\begin{cases}
0.002,& i\in\mathcal{T}_K(t)\\
0.005,& \text{otherwise},
\end{cases}
\qquad K=2.
\]

The top-\(K\) set \(\mathcal{T}_K(t)\) comprises the \(K\) objects whose centroids are closest to the TCP predicted \(\tau=0.3\,\mathrm{s}\) ahead, \(\hat{\mathbf{q}}(t)=\mathbf{q}(t)+\dot{\mathbf{q}}(t)\tau +\tfrac12\ddot{\mathbf{q}}(t)\tau^2\).
The object that is currently nearest to the TCP is always included in the set.

\noindent \paragraph{Behavioural interpretation.} Equations \eqref{eq:cyl_dist}-\eqref{eq:update_rule} implement three intuitive rules: (i) Near, rapidly approaching objects quickly gain probability, mirroring human commitment during the reach phase; (ii) Visually insignificant items (\(p_i(0)<0.05\)) can never exceed \(0.30\), preventing late false positives; (iii) A mild goal bias preserves the most plausible targets through small corrective hand motions.

\begin{figure*}[hb]
    \centering
    \includegraphics[width=1.00\linewidth]{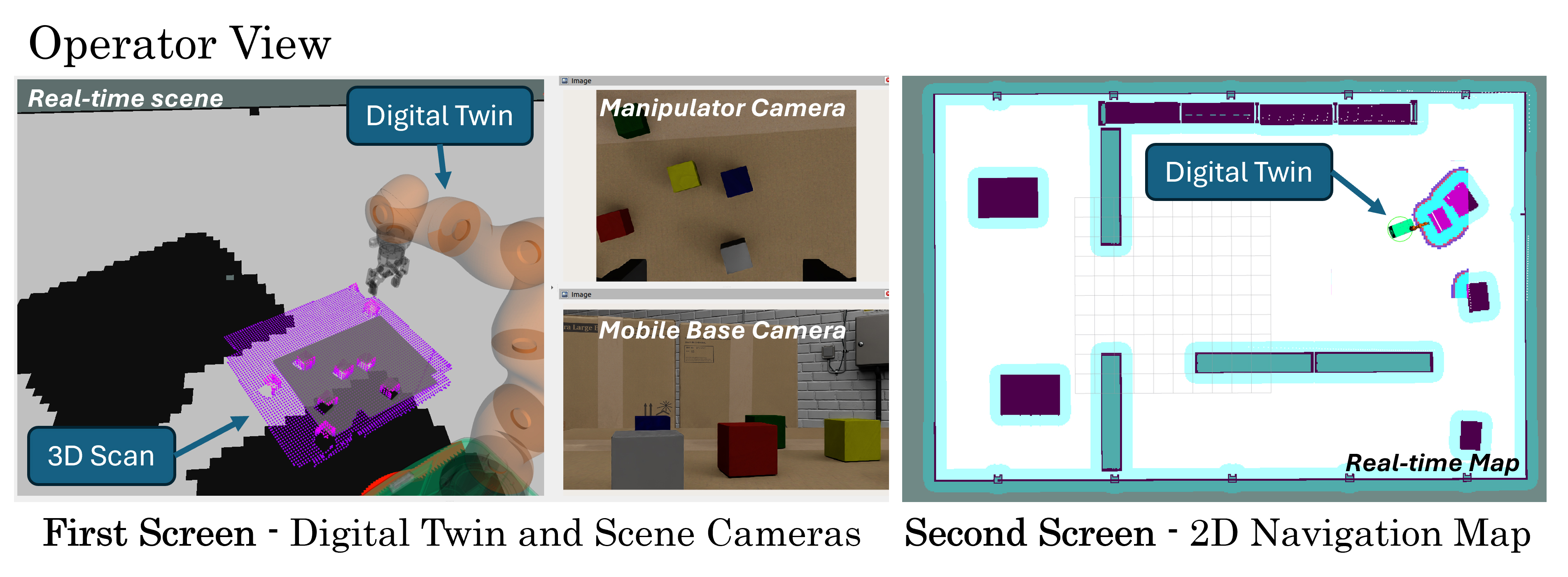}
    \caption{The multi-view interface provided to the operator during teleoperation. The display features a primary 3D view that includes a digital twin and live scan data, as well as camera feeds from the manipulator and mobile base, and a top-down 2D navigation map.}
    \label{fig:map_view}
\end{figure*}

\section{Experimental Evaluation}
\label{sec:experimental-eval}
Experiments were conducted on the high-fidelity simulator Isaac Sim, chosen because of its proven track record of Sim2real transfer without additional tuning \cite{tang2023industreal, albardaner2024sim, salimpour2025sim}. The setup used a simulated Kuka KMR IIWA equipped with a Robotiq 2F-85 gripper, and a Realsense d435i RGB-D camera. The simulation was executed on a computer with Ubuntu 22.04 (RTX 4080 laptop GPU, Intel i7, 64 GB of RAM). The environment, based on Isaac Sim's warehouse setting, features arrangements of shelves, tables, and a diverse array of objects. Participants controlled the robot using a PS4 controller, having access to four views consisting of: (i) A live camera feed from the manipulator, (ii) a front facing base camera, (iii) an isometric digital twin view with a live point cloud overlay, and (iv) a top-down occupancy map. This array is shown in Figure \ref{fig:map_view}.
The mobile base remained omnidirectional throughout both phases. During the navigation phase, the operator moved only joints 1 and 6; the remaining joints were held fixed to simplify control, letting the wrist camera tilt (joint 1) and pan via base rotation (joint 6). During the scanning routine, the robot's motion was fully autonomous, although operators retained the option to interrupt the autonomous sequence at any time. In the manipulation phase, operators could freely position the robot base, while the manipulator accepted Cartesian end-effector targets. A button press triggered autonomous execution to prevent accidental collisions. All operator commands, joint poses, base poses, desired Cartesian positions and trajectories, and sensor data were recorded. No intent-based assistance was provided beyond the autonomous scan and move-to-target actions described above.

\subsection{Experimental Tasks}
Five participants were recruited under ethical approval from the University of Birmingham ethics committee (project ERN\_3337). Each participant performed five distinct trials designed to elicit varied intent patterns. Each trial had a maximum duration of 15 minutes and comprised a navigation phase followed by a manipulation phase once the navigation target was reached. Trials not completed within the time limit were marked as unsuccessful. In total, $5$ participants $\times$ $5$ trials yielded $25$ experiments. The map of the trial with the areas of interest is shown in Figure \ref{fig:map_areas_interest}. The five trial types are defined as follows:

\begin{figure*}[!ht]
    \centering
    \includegraphics[width=1.00\linewidth]{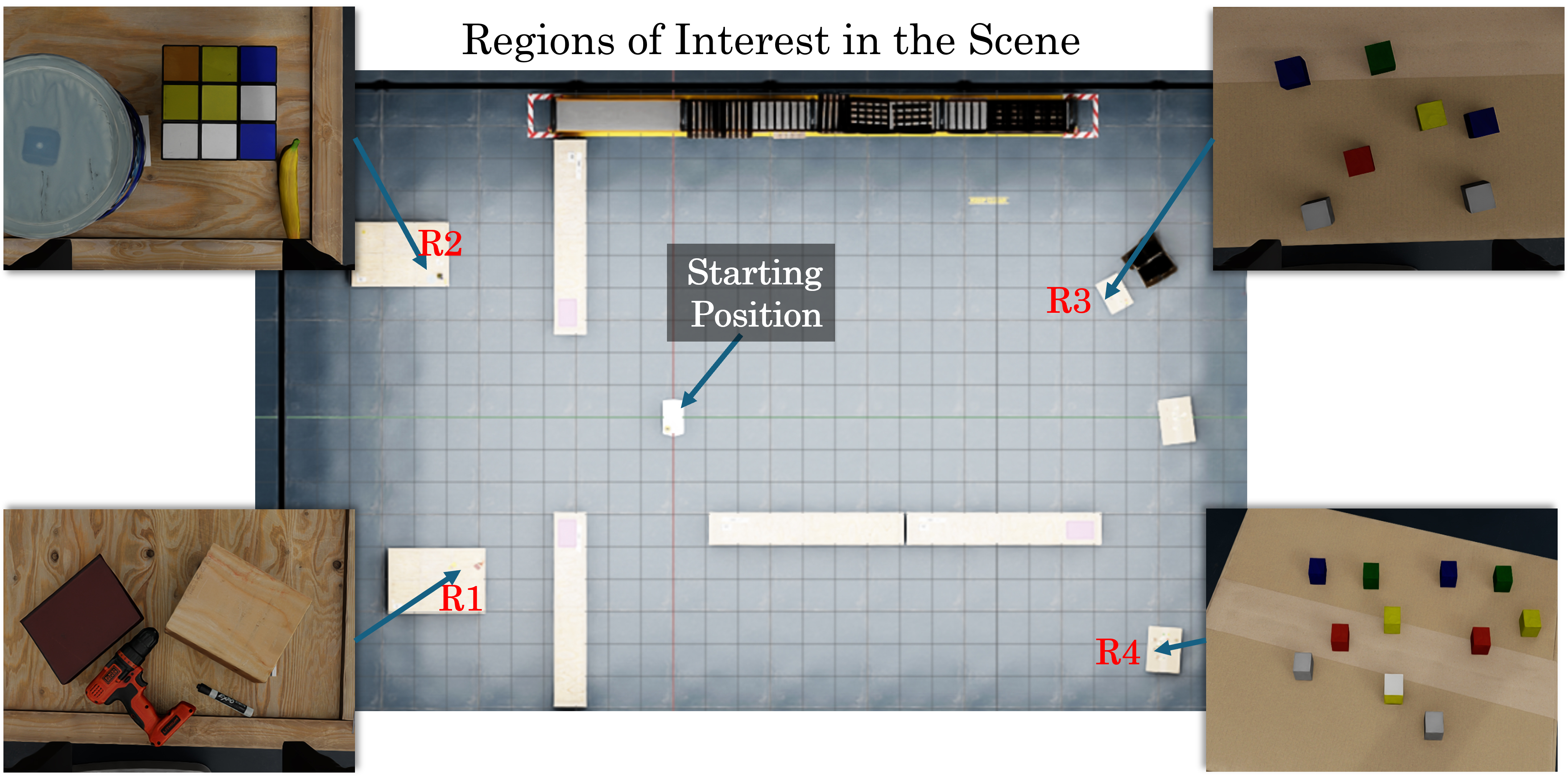}
     \caption{The layout of the experimental environment, highlighting the four primary regions of interest (R1–R4) for the different trials. The zoomed-in images show detailed object arrangements within each region.}
    \label{fig:map_areas_interest}
\end{figure*}

\textbf{T1. Direct Grasp - R3, Yellow Cube.} Drive to region R3 and immediately grasp the Yellow Cube. \emph{Case with a single, unambiguous goal.}

\textbf{T2. Base Redirection - R4$\!\rightarrow$R3, Yellow Cube.} (i) Navigate to R4.\ (ii) Upon arrival, receive a verbal cue redirecting the operator to R3.\ (iii) Grasp the Yellow Cube.\ \emph{Tests redirection at the navigation layer.}

\textbf{T3. Manipulator Redirection - R3.} (i) Navigate to R3.\ (ii) Reach a first instructed cube (colour depends on participant index).\ (iii) Receive a second cue specifying a different cube in the same location (e.g.\ Yellow$\!\rightarrow$Left-Blue).\ (iv) Move the end-effector to the new target and grasp.\ \emph{Tests late redirection in the manipulation layer.}

\textbf{T4. Semantic Tool Grasp - R1, Drill.} Drive to R1 (tool pallet) and identify, then grasp, the electrically powered tool (Drill, without disclosing its name).\ \emph{Checks generalisation of different objects.}

\textbf{T5. Base Redirection \& Manipulator Redirection due to Infeasible Object - R1$\!\rightarrow$R2, coffee can$\!\rightarrow$fruit} (i) Navigate to R1.\ (ii) Upon reaching, receive a cue redirecting to R2.\ (iii) Try to grasp the large coffee can (end-effector too small).\ (iv) Receive a final cue to grasp the adjacent fruit.\ \emph{Stresses base redirection, manipulator redirection, and handling of infeasible geometric actions.}

\noindent (All experiments share the same initial position at the center of the map.)

\subsection{Evaluation Metrics}
We assess GUIDER with two metrics: 

\textbf{Remaining Time at Confident Prediction (RTCP)}: The time interval from the moment the correct target first achieves and maintains (for at least 0.5\,s) the highest predicted probability, until actual contact (navigation) or operator command (manipulation). Higher RTCP indicates earlier intent recognition.

\textbf{Prediction Stability}: The percentage of the remaining task duration (after the first correct prediction) during which the predicted intent remains correctly focused on the intended target. Higher stability implies sustained correctness.
\\

\noindent Predictions are compared against manually annotated ground-truth labels generated offline from the known task instructions. We benchmark our method against two methods: (i) Robot Trajectron \cite{song2024robot}, a state-of-the-art manipulator-arm trajectory predictor, and (ii) Bayesian Operator Intent Recognition (BOIR) \cite{panagopoulos2021bayesian}, our previous navigation intent-focused method. We pass the recorded teleoperation data through all methods. Every base pose, joint trajectory data, and the RGB-D frame are fed to GUIDER, Robot Trajectron, and BOIR.  Because the baselines need explicit goals, we provide BOIR with the four navigation regions (R1–R4) and Robot Trajectron with the object list returned by our perception module. Each algorithm outputs an intent estimate at every time step, which we score with the RTCP and stability metrics.

As the initial hypothesis, higher stability is expected in both phases, expressed as $\mathrm{Stab}_{\text{GUIDER}} > \mathrm{Stab}_{\text{baseline}}$, because the growth-decay hysteresis damps sudden goal flips. No overall change in reaction time is anticipated $\bigl(\mathrm{RTCP}_{\text{GUIDER}} \approx \mathrm{RTCP}_{\text{baseline}}\bigr)$. Initial predictions from the perception module in manipulation should give a good first approximation, but are expected only to have a substantial effect in geometrically constraint cases, where the feasibility check should yield a much earlier prediction, $\mathrm{RTCP}_{\text{GUIDER}} < \mathrm{RTCP}_{\text{baseline}}$. A statistical analysis was executed that, for each case, summarises the five paired observations with the median ± MAD and then applies the exact paired Wilcoxon signed-rank test. The two-tailed $p_{2}$ gauges any difference, while the one-tailed $p_{1}$ tests the a priori hypothesis. The matched-pairs rank-biserial correlation $r_{\text{bs}}$ accompanies the $p$-values to convey effect magnitude, completing an evaluation reported as $p_{2}/p_{1}/r_{\text{bs}}$ alongside each median comparison.

\begin{table*}[hb]
  \centering\footnotesize
  \caption{Paired intent prediction results.}
  \label{tab:intent_stats_robust}

  \begin{threeparttable}
  \setlength{\tabcolsep}{4pt} 
  \begin{tabularx}{\textwidth}{@{}L| N N Y|N N Y@{}}
  \toprule
  \makecell[c]{\textbf{Navigation Phase}} &
  \multicolumn{3}{c}{\makecell[c]{RTCP [s] \\ (Median ± MAD, higher = earlier)}} &
  \multicolumn{3}{c}{\makecell[c]{Stability [\%] \\ (Median ± MAD, higher = steadier)}}\\
  \cmidrule(lr){2-4}\cmidrule(lr){5-7}
  Case &
  GUIDER & BOIR~\cite{panagopoulos2021bayesian} & $p_{2}/p_{1}/r_{\text{bs}}$ &
  GUIDER & BOIR & $p_{2}/p_{1}/r_{\text{bs}}$\\
  \midrule \midrule
  T1 Direct \\(R3)& 56.4±0.5 & 63.9±2.5 & .312/.156/-0.60 & 95.1±2.8 & 87.0±2.8 & .312/.156/+0.60\\
  \dottedmidrule
  T2 Base Redirection \\(R4$\!\rightarrow$R3)& 63.6±10.1 & 59.8±6.4 & .062/.031/+1.00 & 100.0±0.0 & 71.3±16.0 & .125/.062/+1.00\\
  \dottedmidrule
  T3 Manip. Redirection \\(R3)& 69.3±3.3 & 71.2±4.1 & .125/.062/-0.87 & 92.9±0.6 &100.0±0.0& .062/.031/-1.00\\
  \dottedmidrule
  T4 Tool grasp \\(R2)& 41.5±9.8 & 42.9±9.6 & .062/.031/-1.00 &100.0±0.0& 97.7±2.3 & .250/.125/+1.00\\
  \dottedmidrule
  \textbf{T5 Base Redirect} \\ + \textbf{Geometric Redirect}\\(R1$\!\rightarrow$R2)&
  40.6±5.6 & \textcolor{red}{74.0±5.9\textsuperscript{\dag}} & - &
  100.0±0.0 & 60.5±4.4 & \textbf{.062/.031/+1.00}\\
  \midrule
  \textit{\textbf{Pooled Stability (n=25)}} & - & - & - &
  100.0±0.0 & 87.0±13.0 & \textbf{.006/.003/+0.68}\\
  \midrule[1.2pt]

  \multicolumn{1}{c}{\textbf{Manipulation Phase}} &
  \multicolumn{3}{c}{RTCP [s]} &
  \multicolumn{3}{c}{Stability [\%]}\\
  \cmidrule(lr){2-4}\cmidrule(lr){5-7}
  Case &
  GUIDER & Trajectron~\cite{song2024robot} & $p_{2}/p_{1}/r_{\text{bs}}$ &
  GUIDER & Trajectron & $p_{2}/p_{1}/r_{\text{bs}}$\\
  \midrule
  \midrule
  T1 Direct \\(Cube A) & 18.8±1.8 & 17.5±4.6 & .125/.062/+0.87 & 93.9±6.1 & 97.9±2.1 & .875/.438/-0.20\\
  \dottedmidrule
  T2 Base Redirection \\(Cube A) & 12.0±4.0 &  9.5±1.3 & .125/.062/+0.87 &100.0±0.0 &100.0±0.0 & -\\
  \dottedmidrule
  T3 Manip. Redirection \\(Cube A$\!\rightarrow$Cube B) &  4.0±1.8 &  6.2±4.6 & .188/.094/-0.73 &100.0±0.0 & 68.6±31.5 & .625/.312/+0.30\\
  \dottedmidrule
  T4 Tool grasp \\(Drill) & 12.2±1.2 & 13.3±2.0 & .625/.312/-0.33 &\textbf{100.0±0.0} & 87.2±6.6 & \textbf{.062/.031/+1.00}\\
  \dottedmidrule
  \textbf{T5 Base Redirect} \\+ \textbf{Geometric Redirect}\\ (Can$\!\rightarrow$Fruit) &
  \textbf{23.6±3.5} &  7.8±3.6 & \textbf{.062/.031/+1.00} &
  100.0±0.0 &100.0±0.0 & -\\
  \midrule
  \textit{Pooled Stability (n=25)} & - & - & – &100.0 ± 0.0 &100.0 ± 0.0 & .109/.054/+0.38\\
  \bottomrule[1.2pt]
  \end{tabularx}

  \begin{tablenotes}\footnotesize
  \item[] $p_{2}$: two-tailed exact Wilcoxon; $p_{1}$: one-tailed, pre-registered; $r_{\text{bs}}$: rank-biserial correlation.
  \item[] \textcolor{red}{\(\dagger\)} BOIR first locked onto region R1, so the inflated RTCP reflects an incorrect early prediction.
  \item[] Bold marks directional results with \(p_{1}<0.05\) and perfect or large effect (\(|r_{\text{bs}}|\ge 1\)).
  \end{tablenotes}

  \end{threeparttable}
\end{table*}

Although the setup allows for the simultaneous control of base and manipulator, performance comparisons are conducted separately for the navigation and manipulation phases, as to the best of our knowledge, no other framework exists that deals with predictions during the same task involving both the mobile base and manipulator simultaneously.

\section{Results} \label{sec:results}
We evaluate GUIDER against baseline methods in both navigation and manipulation phases across five distinct task scenarios. The performance was assessed using the RTCP, where higher values indicate earlier correct predictions, and Prediction Stability, with higher values indicating a more consistent prediction throughout the test. The results for this evaluation are reported in Table \ref{tab:intent_stats_robust}.

For the navigation phase, our approach achieved a substantial improvement in prediction stability compared to the BOIR baseline, as previously hypothesised. Median stability reached 100\% across the pool navigation trials, significantly higher than BOIR's overall median stability of 87.0\%. Stability improvements were most pronounced during tasks requiring redirection (T2 and T5), where GUIDER maintained a stability of 100\% for most tasks, whereas BOIR dropped to 71.3\% (T2) and 60.5\% (T5).

Regarding RTCP, our method showed mixed outcomes. Generally, it matched BOIR’s performance across tasks, or was a few seconds slower. However, in task T5, BOIR prematurely locked onto an incorrect region, resulting in an inflated RTCP (74.0 s), while our method provided a more realistic and accurate median RTCP (40.6 s). Although both methods managed to predict the correct target at some point during the 25 trials, GUIDER maintained the correct prediction until contact with the target 25 out of 25 times, while BOIR failed in 1 out of the 25 trials.

In the manipulation phase, GUIDER consistently achieved a high median stability (100\%) across all tasks, indicating robust and sustained predictions throughout manipulation activities. In contrast, Trajectron displayed varying stability, performing well (100\%) on some scenarios (T2 and T5), but worse during the other scenarios (T3, 68.6\%) and tool grasp task (T4, 87.2\%).

Regarding prediction timing (RTCP), our method produced comparable results to Trajectron for most tasks, with no statistically significant differences. Notably, our method significantly improved RTCP in the geometrically constrained manipulation scenario (T5), achieving a median of 23.6 s, compared to Trajectron's 7.8 s, as previously hypothesized due to our integrated geometric feasibility constraints, which promptly identified infeasible grasp targets. GUIDER managed to keep correct predictions at the moment of contact with the target object in 96\% of the trials, while Trajectron maintained it in 92\% of the cases.

\begin{figure*}[h]
    \centering
    \includegraphics[width=1.00\linewidth]{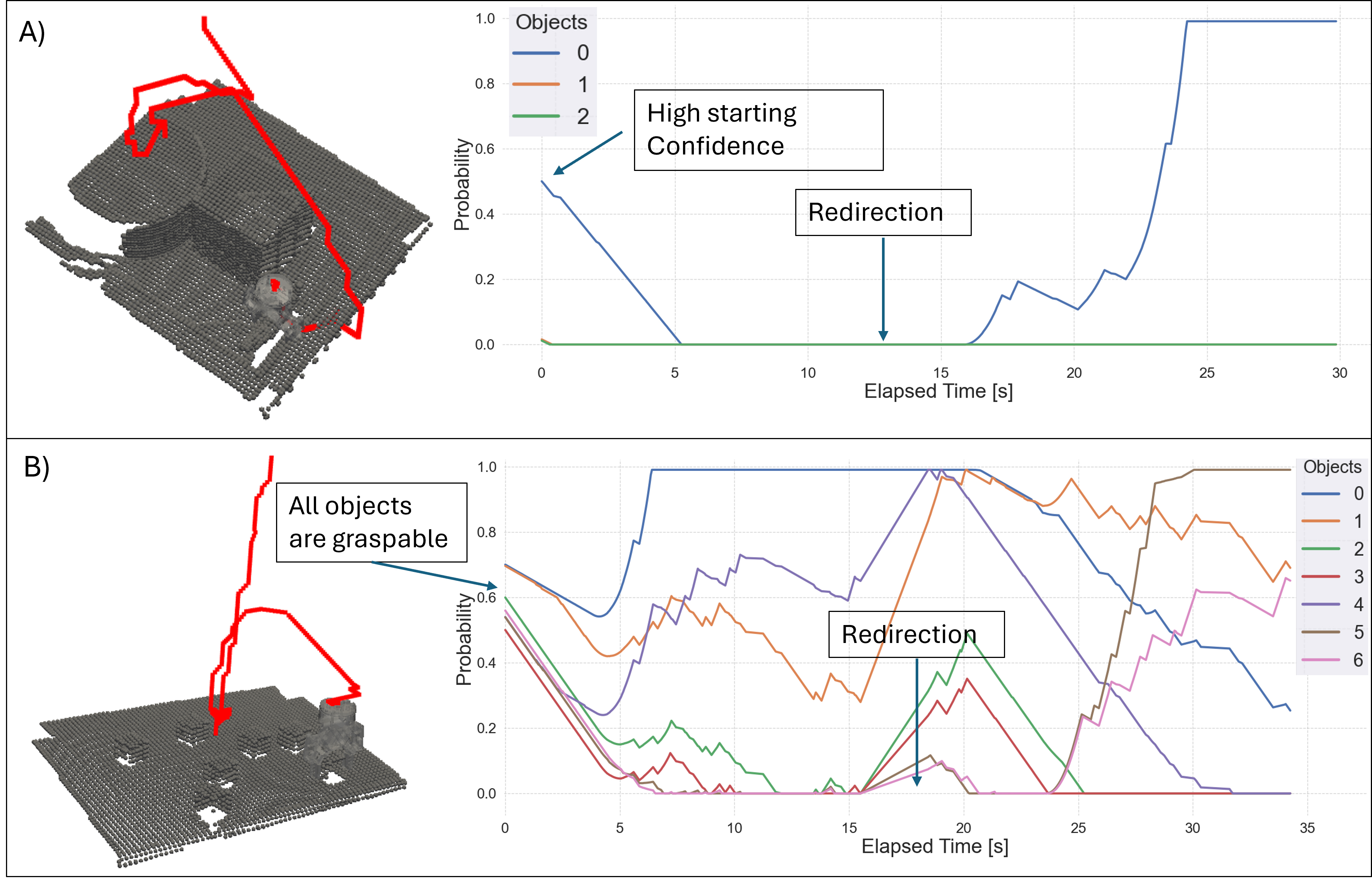}
    \caption{Example plots of the end-effector probability evolution. The y-axis represents the probability of each object being the intended target, while the x-axis represents time. A) A Geometric Redirection Trial (T5), showing how the geometric constraint quickly shifts its focus to the feasible object; B) A Manipulator Redirection trial (T3) illustrating how the system's belief shifts from one object to another in response to the operator's actions.}
    \label{fig:prob_evolution_example}
\end{figure*}

Figure \ref{fig:prob_evolution_example} presents representative probability-evolution plots for redirection trials T3 and T5. In both trials, probabilities assigned to unlikely objects decay over time; after a redirection, GUIDER shows a steady increase for the correct object. During the manipulation stage, the perceptual module provides an initial ranking. In the geometrically constrained scenario (T5), GUIDER ranked the food object first in all five trials. Across the non-redirection tasks (T1, T2, T4), the target object maintained a median rank of 2.

\section{Discussion}
\label{sec:discussion}
The experimental results show several strengths and considerations of our method. The distinction between navigation and manipulation phases, utilizing different primary information sources, appears necessary for handling the various stages of mobile manipulation tasks. But embedding them together provides the advantage of being able to work in a larger workspace, which requires shifts between both modes.

For the navigation phase GUIDER substantially improved prediction stability by integrating spatially-informed motion evidence with map context, particularly noticeable during redirected navigation tasks (T2, T5). The Synergy Map facilitated smooth belief adjustments that grew and decayed gradually, preventing transient controller motions from causing rapid prediction shifts, unlike the BOIR baseline.

Similarly, our manipulation phase consistently saw high stability values throughout the tasks due to the integration of perceptual saliency and geometric feasibility checks in the case of T5, which had geometric constraints. For the other cases, the growth and decay function, as well as the velocity and acceleration-aware probabilistic evolution function of the end-effector, played a role in maintaining hysteresis and locking of targets. In contrast, Trajectron exhibits extreme sensitivity to small and sudden rapid shifts.

Overall, our method improves on prior intent-inference methods in three important qualitative ways among other improvements. (a) Goal-free inference: relevant objects and areas emerge during operation, eliminating the need for predefined waypoints or object lists. (b) Coupled probabilistic backbone spanning base and arm: a unified belief supports both phases, enabling future variable-autonomy controllers to take over navigation or manipulation tasks seamlessly once intent confidence is sufficient. (c) Geometry-aware object ranking: The geometric feasibility cascade down-weights objects that the gripper cannot physically manipulate. Only feature (c) influenced quantitative scores in our offline study; the other features enabled automatic identification of intent-relevant areas/objects and smooth redirection handling for both base and manipulator movements.

\paragraph{Limitations and Future Work}
Our method and testing present some limitations with this study serving as a proof-of-concept for GUIDER, with the limitations expected to be addressed in future work. First, the evaluation is currently limited to simulation data collected from five participants. However, statistically significant and robust effect sizes emerged, as predicted by the hypothesis we set out to prove. A second limitation is that hyperparameters were manually tuned on a preliminary experiment (not included in the reported trials) and kept fixed for all experiments; adaptive parameter learning methods, such as Bayesian optimization \cite{victoria2021automatic} or reinforcement learning \cite{sehgal2019deep}, could enhance the framework’s adaptability and performance across varying tasks and conditions, for this a broader set of tests and conditions, and a wider number of participants or emulated trajectories based on human styles \cite{pmlr-v229-mandlekar23a} is required.  A third limitation lies in the vision module, rather than directly with GUIDER. The RGB camera detected every real object but occasionally produced extra masks under shadows or in areas with glare. The fusion stages assign each mask a separate probability, so these false positives receive low confidence and rarely affect intent ranking. Integrating multi-view or full 3D perception would further reduce false positives and accelerate inference.

Future work with GUIDER will validate the pipeline in closed-loop real-world experiments, subjecting the framework to more realistic sensor noise, latency, and occlusions. Additionally, we aim to provide real-time operator assistance and adjustable autonomy responsive to inferred intent. Different operator control modalities and types of robots can also be tested, alongside richer spatial feedback techniques to enhance operator situational awareness \cite{rastegarpanah2024electric,mineo2023edge,rastegarpanah2025haptic,rastegarpanah2024semi}. Furthermore, we plan to augment the perception module with full 3D saliency based on methods such as Local Contact Matching or spectral correlation \cite{adjigble2018model,adjigble2021spectgrasp}. Finally, integration of multimodal reasoning and world models will be explored to enable inference of complex tasks and long-term user goals \cite{team2025gemma,team2025gemini}.

\section{Conclusion}
\label{sec:conclusion}
This paper introduced GUIDER, a dual-phase probabilistic framework designed to unify navigation and manipulation intent inference for human-robot collaboration.  By creating a unified belief structure that operates without predefined goals, GUIDER addresses a limitation of current systems that treat these two essential phases of mobile manipulation as isolated problems.  The framework’s strength lies in its human-inspired approach, which first reasons about high-level spatial goals via the combined belief, or synergy map, and then shifts focus to fine object interaction by integrating perceptual saliency, instance segmentation, and multiple geometric feasibility checks.

Experiments showed higher prediction stability for GUIDER compared to the baselines, in both phases.  This robustness stems from a design that evolves its beliefs, utilizing motion context, hysteresis, visual saliency, and geometric constraints to help avoid erratic predictions. In geometry-constrained tasks, GUIDER recognized the correct, feasible object approximately 3 times sooner than the baseline, a direct result of its built-in grasp affordance analysis.

Saliency-based object ranking works exceptionally well in scenarios where one object is visually dominant, and all the prior assumptions from perception are met. Still, in scenarios with multiple similar objects, the end-effector motion is the decisive factor in disambiguation. This indicates our multi-step approach in GUIDER effectively combines visual and operator cues. While our framework can occasionally assign some probability to an incorrect object early on, the design ensures that unless an object meets all criteria (saliency, feasibility, and approach consistency), it will not maintain a dominating probability. In a real-world setting, an assistive controller could use a confidence threshold before intervening.

A single probabilistic backbone for both base and arm lays some groundwork for variable autonomy controllers that can hand over manoeuvring or grasp execution when operator intent is sufficiently certain. 

\backmatter

\section*{Declarations}

\subsection*{Funding}
This work was funded by the Nuclear Decommissioning Authority (NDA) and supported by the United Kingdom National Nuclear Laboratory (UKNNL). In addition, it was supported by the UK Research and Innovation (UKRI) project “REBELION" under Grant 101104241.

\subsection*{Conflict of interest}
The authors declare that they have no competing interests.  

\subsection*{Ethics approval}
The study protocol, including all operator-in-the-loop experiments, was approved by the University Ethics Committee under project ERN\_3337.

\subsection*{Consent for publication}
All authors reviewed the manuscript and agreed to its publication.  

\subsection*{Data availability}
The datasets generated and analysed during the current study are available from the corresponding author upon reasonable request.

\subsection*{Materials availability}
Not applicable.

\subsection*{Code availability}
The source code will be released later and can be obtained from the corresponding author upon reasonable request for non-commercial research purposes.

\subsection*{Author contribution}
\textbf{C.A.C.}: Conceptualisation; Methodology; Software; Formal analysis; Investigation; Writing - original draft. \textbf{M. C.}: Supervision; Resources; Writing - review \& editing. \textbf{A. R.}: Validation; Visualization; Writing - review \& editing; Project administration; Funding acquisition; Supervision. \textbf{M. S.}: Industrial guidance; Resources; Writing - review \& editing.
\textbf{R. S.}: Project administration; Funding acquisition; Supervision; Writing - review \& editing.

\bibliography{sn-bibliography.bib}

\end{document}